\documentclass{article}
\PassOptionsToPackage{numbers, compress}{natbib}
\usepackage{multirow}
\usepackage{amsmath}
\DeclareMathOperator{\argmax}{argmax}

\usepackage{algorithm}
\usepackage{algpseudocode}  
\usepackage{soul}    

\usepackage{xcolor}  
\definecolor{darkgreen}{RGB}{0,128,0}
\newcommand{\posval}[1]{{\textcolor{darkgreen}{#1}}}
\newcommand{\negval}[1]{{\textcolor{red}{#1}}}
\usepackage{url}
\definecolor{citeblue}{rgb}{0.21,0.49,0.74}
\usepackage[pagebackref=false,breaklinks,colorlinks,citecolor=citeblue,bookmarks=false]{hyperref}
\usepackage[preprint]{neurips_2026}

\usepackage[utf8]{inputenc} 
\usepackage[T1]{fontenc}    
\usepackage{hyperref}       
\usepackage{url}            
\usepackage{booktabs}       
\usepackage{amsfonts}       
\usepackage{nicefrac}       
\usepackage{microtype}      
\usepackage{xcolor}         
\usepackage{eso-pic}       
\usepackage{graphicx}      
\usepackage{calc}          
\usepackage{colortbl}
\usepackage{xspace}
\newcolumntype{C}{>{\centering\arraybackslash}p{0.04\textwidth}}
\newcommand{\methodname}{\textsc{Skill0}\xspace}
\definecolor{HeaderGray}{RGB}{245,245,245}
\definecolor{BlockGray}{RGB}{250,250,250}
\definecolor{OursBlue}{RGB}{234,244,255}
\definecolor{TokenOrange}{RGB}{255,245,235}
\definecolor{DeepRed}{RGB}{165,42,42}
\definecolor{DeepPurple}{RGB}{63,0,95} 
\definecolor{DeepGreen}{RGB}{0,160,80}
\definecolor{DeepBlue}{RGB}{30,60,180}
\definecolor{brown}{RGB}{139,69,19}
\definecolor{DeepOrange}{RGB}{204,85,0}
\definecolor{thirdcolor}{RGB}{206, 231, 196}
\definecolor{topcolor}{RGB}{252, 236, 196}
\definecolor{secondcolor}{RGB}{223, 235, 253}
\newcommand{\mycomment}[1]{\textcolor{darkgreen}{\textit{// #1}}}
\usepackage[most,skins,theorems]{tcolorbox}

\newtcolorbox{templatebox}[1]{
  enhanced,
  unbreakable,
  colback=white,
  colframe=black!65,
  colbacktitle=black!80,
  coltitle=white,
  boxrule=0.9pt,
  arc=2pt,
  left=6pt,
  right=6pt,
  top=6pt,
  bottom=6pt,
  title={#1},
  fonttitle=\bfseries,
  sharp corners,
  boxed title style={sharp corners, boxrule=0pt}
}
\usepackage{amsthm}

\usepackage{amsmath}
\usepackage{amssymb}

\usepackage{thmtools}
\theoremstyle{plain}

\theoremstyle{definition}

\theoremstyle{remark}

\AddToShipoutPictureBG{%
  \ifnum\value{page}=1
    \AtPageUpperLeft{%
      \put(\LenToUnit{3.8cm},\LenToUnit{-2.7cm}){%
        \includegraphics[height=0.8cm]{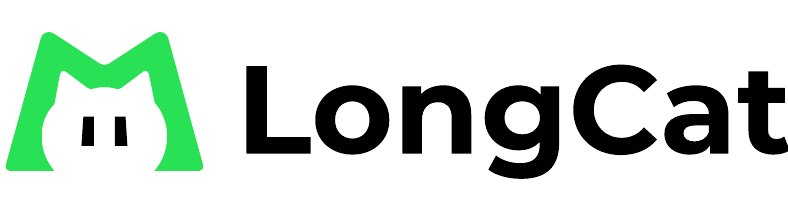}%
      }%
    }%
  \fi
}
\title{\methodname: In-Context Agentic Reinforcement \\ Learning for Skill Internalization}

\author{
\textbf{Zhengxi Lu}$^{1,2\star}$, \textbf{Zhiyuan Yao}$^{2}$, 
\textbf{Jinyang Wu}$^{3}$, \textbf{Chengcheng Han}$^{2}$, \textbf{Qi Gu}$^{2\dagger}$\\
\textbf{Xunliang Cai}$^{2}$,
\textbf{Weiming Lu}$^{1}$, 
\textbf{Jun Xiao}$^{1}$,
\textbf{Yueting Zhuang}$^{1}$,
 \textbf{Yongliang Shen}$^{1\dagger}$\\[3pt]
  $^1$Zhejiang University \qquad$^2$Meituan \qquad$^3$Tsinghua University\\
  \texttt{\small \{zhengxilu, syl\}@zju.edu.cn\qquad guqi03@meituan.com} \\
  \begin{tabular}{@{}ll@{}}
  \end{tabular}}

\begin{document}

\maketitle
\vspace{-16pt}
\begin{abstract}

Agent skills, structured packages of procedural knowledge and executable resources that agents dynamically load at inference time, have become a reliable mechanism for augmenting LLM agents. 
Yet \textit{inference-time skill augmentation} is fundamentally limited: retrieval noise introduces irrelevant guidance, injected skill content imposes substantial token overhead, and the model never truly acquires the knowledge it merely follows. 
We ask whether skills can instead be internalized into model parameters, enabling zero-shot autonomous behavior without any runtime skill retrieval.
We introduce \textbf{\methodname{}}, an in-context reinforcement learning framework designed for \textit{skill internalization}.
\methodname{} introduces a training-time curriculum that begins with full skill context and progressively withdraws it. Skills are grouped offline by category and rendered with interaction history into a compact visual context, teaching the model tool invocation and multi-turn task completion.
A \textbf{Dynamic Curriculum} then evaluates each skill file's on-policy \textit{helpfulness}, retaining only those from which the current policy still benefits within a linearly decaying budget, until the agent operates in a fully zero-shot setting.
Extensive agentic experiments demonstrate that \methodname{} achieves substantial improvements over the standard RL baseline (+9.7\% for ALFWorld, +6.6\% for Search-QA and +10.1\% for WebShop), while maintaining a highly efficient context of fewer than 0.5k tokens per step. Our code is available at \url{https://github.com/ZJU-REAL/SkillZero}.

\end{abstract}

\section{Introduction}

\begin{figure}[h]
\centering
\begin{minipage}{0.46\columnwidth}
\begin{center}
    \textit{``\textbf{Skills} at training, \textbf{zero} at inference.''}
\end{center}
\begin{flushright}
    \textit{--- \methodname{}}
\end{flushright}
    Large Language Models (LLMs)~\citep{guo2025ds-r1,team2025kimi,yang2025qwen3,comanici2025gemini,team2026longcat-2601} have demonstrated strong decision-making capabilities across complex real-world tasks, including code generation~\citep{jimenez2023swebench}, 
GUI automation~\citep{ye2025mobileagentv3}, 
gameplay~\citep{shridhar2020alfworld}, and embodied control~\citep{wang2023voyager}.
 With the emergence of agent scaffolds like Claude Code and OpenClaw, structured Agent Skills~\citep{xu2026agentskillsurvey,li2026skillecosystem,he2026openclaw}, which are defined as compact and reusable strategies that capture the principles, have become the standard mechanism for extending agent capabilities on specialized tasks.
\end{minipage}
\hfill
\begin{minipage}{0.5\columnwidth}
    \centering
    \includegraphics[width=\linewidth]{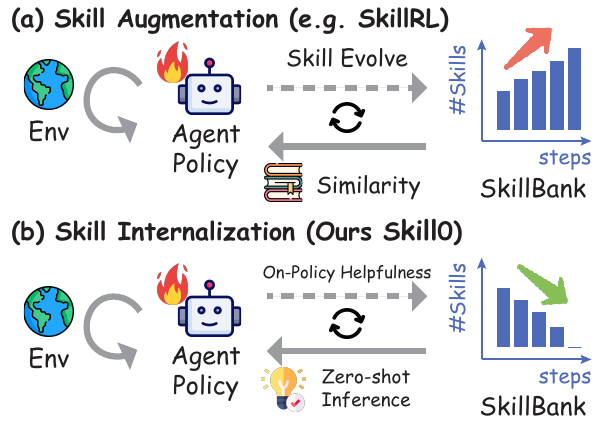}
    \caption{Comparison of \textbf{(a)} Skill Augmentation methods and \textbf{(b)} our Skill Internalization method.}
    \label{fig:motivation}
\end{minipage}

\end{figure}
\vspace{-5pt}

\newpage

The prevailing paradigm is inference-time skill augmentation: relevant skills are retrieved from a skill bank and injected into the model's context as natural language guidance at each step~\citep{li2026skillsbench,liang2026skillnet}. This approach has proven effective and is now well-established, with growing ecosystems of skill libraries, retrieval pipelines, and evolution mechanisms~\citep{xu2026agentskillsurvey,xia2026skillrl}.
Yet this practical success obscures a more fundamental limitation.
First, retrieval noise introduces irrelevant or misleading guidance that corrupts the agent's context~\citep{yao2026toolace,wang2026agentnoisebench}.
Second, injected skill content imposes token overhead that compounds across multi-turn interactions, limiting scalability~\citep{liu2024lost,hsieh2024ruler}.
Third, and most critically, a model that follows skill descriptions in its prompt is executing skills, not learning them: competence resides in the context, not in the model~\citep{wang2025rlskill,han2026swe}.

This observation suggests a different question: rather than asking how to better retrieve and inject skills, can skills be \textit{internalized} into model parameters, rendering retrieval unnecessary at inference time? 
Skill acquisition in humans follows a familiar progression: an explicit instruction phase gives way to an internalized phase in which the same behavior is executed autonomously from memory~\citep{anderson1982acquisition, yuan2025fxgxfgxllms}.
Inference-time skill augmentation permanently anchors agents in the first stage. 
Reinforcement learning offers a natural path to the second, driving the agent to consolidate effective strategies as intrinsic policy rather than reading them from context~\citep{guo2025deepseekr1,shao2024deepseekmath}. 
Yet a naive application of RL fails in both directions: without skill context, the agent lacks the structured guidance necessary to learn complex multi-step behavior; with full skill context throughout, the model remains dependent on external knowledge it has never been required to internalize. What is needed is a training regime that starts with skills and ends without them, systematically transferring competence from context to parameters.

We propose \textbf{\methodname{}}, the first RL framework that formulates skill internalization as an explicit training objective. \methodname{} realized this curriculum through In-Context Reinforcement Learning (ICRL): skills are provided as in-context guidance during training rollouts but removed entirely at inference, so that RL optimization directly drives the transition from context-dependent execution to autonomous behavior. Concretely, skills are grouped offline by category and rendered with interaction history into a compact visual context, teaching the model tool invocation and multi-turn task completion.
\textbf{Dynamic Curriculum} evaluates each skill file's on-policy helpfulness by comparing agent performance with and without it on a matched validation sub-task. Skills are retained only where the current policy still benefits, and discarded otherwise, until the budget reaches zero and the agent operates without any skill context.
Extensive experiments demonstrate that \methodname{} achieves substantial improvements over strong baselines like AgentOCR ((+9.7\% for ALFWorld, +6.6\% for Search-QA and +10.1\% for WebShop)), and competitive performance against skill-augmented methods like SkillRL. Notably, by eliminating skill reliance at inference time, \methodname{} maintains 
a highly efficient context of fewer than 0.5k tokens per step, 
significantly reducing inference overhead without sacrificing task performance.

\begin{itemize}
  \item We propose \methodname{}, the first RL framework that formulates skill internalization as an explicit training objective, moving agents from inference-time skill dependence to fully autonomous zero-shot behavior.
  \item We introduce {in-context reinforcement learning}, which provides structured skill guidance during training rollouts and removes it entirely at inference, directly optimizing the transition from context-dependent execution to intrinsic competence.
  \item We propose {Dynamic Curriculum}, a helpfulness-driven annealing mechanism that withdraws each skill only when the current policy no longer benefits from it, replacing rigid schedules with adaptive internalization.
\end{itemize}

\section{Related Work}
\subsection{LLM Agents}
Recent advancements in instruction-tuned LLMs have enabled autonomous agents 
to operate across a wide range of dynamic, open-world environments, 
including code generation~\citep{jimenez2023swebench,wang2026codea1}, 
GUI automation~\citep{ye2025mobileagentv3,liu2026memgui}, 
gameplay~\citep{shridhar2020alfworld}, and embodied control~\citep{wang2023voyager}. 
With the recent development of reinforcement learning for LLMs~\citep{yu2025dapo,zheng2025group,yao2026coba,chen2026learning}, agentic RL has emerged as a crucial post-training recipe for equipping LLM agents with robust decision-making capabilities~\citep{lu2026uir1,lu2025uis1,feng2025gigpo}.

\begin{figure}[t]
\centering
\includegraphics[width=\columnwidth]{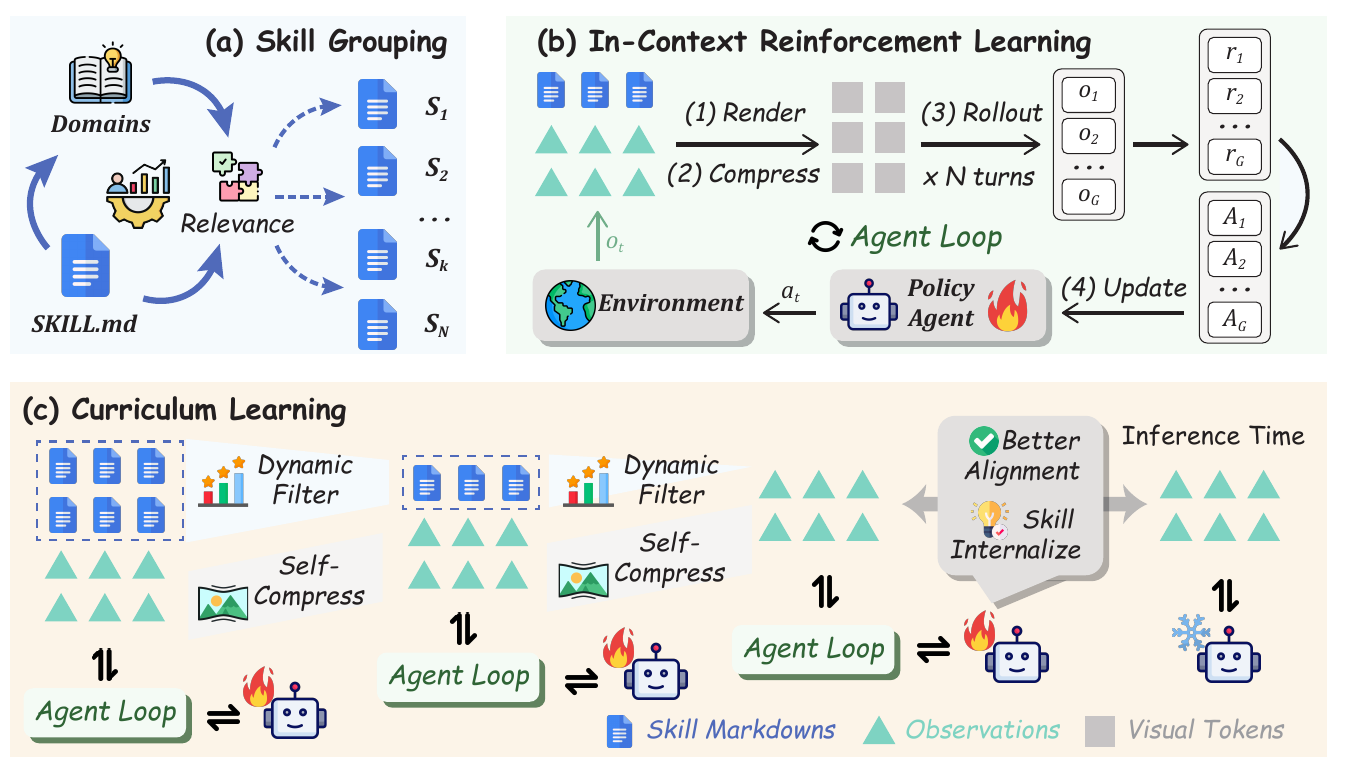}
\caption{\textbf{Overview of \methodname.} (a) Relevance-Driven Skill Grouping; (b) In-Context Reinforcement Learning with skill-enhanced agent loop; (c) Dynamic curriculum learning during training process.}
\label{fig:method}
\end{figure}
\subsection{Agent Skills}
Early memory-based approaches store raw trajectories directly into external databases 
during the sampling process, serving as references for experience replay~\citep{zhao2024expel,shinn2024reflexion}. 
However, such raw trajectories are often lengthy, redundant, and noisy, 
making direct injection into the context window inefficient~\citep{chhikara2025mem0}. 
To address this limitation, a growing line of work has explored \textit{skills}---reusable, 
abstracted, and structured behavioral primitives distilled from historical 
trajectories~\citep{xu2026agentskillsurvey,li2026skillecosystem,he2026openclaw}. 
Skills serve as a form of episodic memory that agents can consult at decision 
time~\citep{li2026skillsbench,liu2026self-vla,liang2026skillnet}, 
and have further been shown to provide efficient guidance within reinforcement 
learning frameworks~\citep{xia2026skillrl,wang2025rlskill,jiao2026agenticproposing}. 
Despite these advances, existing methods predominantly focus on skill extraction, 
organization, and retrieval, leaving the question of whether skills can be 
\textit{internalized} into model parameters largely unexplored.

\section{Method: \methodname}
\label{sec:method}

\subsection{Agent Loop}
\paragraph{Task Definition.} We formulate agent automation as a sequential decision-making problem. Given a task instruction $I$, the agent generates a sequence of actions $\{a_1, a_2, \ldots, a_T\}$ to complete the task. At each step $t$, the agent operates within a structured environment $\mathcal{E}$ (e.g., an online simulator or retrieval engine), which provides a textual observation $o_t$ describing the current environmental state. The agent then samples an action $a_t$ from policy $\pi_{\theta}(a_t | I, h_{t})$, where $\theta$ denotes model parameters and $h_{t}$ represents history up to time $t$. 
\begin{equation}
    h_t=\{o_1, o_2, \ldots, o_t\}
\end{equation}
The environment $\mathcal{E}$ transitions to
the next state and returns the next observation $o_{t+1} = \mathcal{E}(o_t, a_t)$. This rollout continues until the task is successfully completed or the max step threshold is reached.
\paragraph{Skill Management.}
Following~\citet{xia2026skillrl}, we organize reusable behavioral knowledge 
into a hierarchical skill library \texttt{SkillBank}, structured into two levels:
\textbf{(1) General skills} capture universal strategic principles 
applicable across all task types, such as exploration strategies and 
goal-tracking heuristics.
\textbf{(2) Task-specific skills} store specialized knowledge 
for task category $k$, including domain-specific action sequences and preconditions.

Skills are organized in a directory structure 
\texttt{``skills/\{task\_name\}/\{skill\_category\}.md''}, 
where each Markdown file $\mathcal{S}_k$ stores a group of related skills 
sharing the same task and skill category 
(e.g., \texttt{``skills/search/entity\_attribute\_lookup.md''}).
The complete library $\texttt{SkillBank} = \{\mathcal{S}_k\}_{k=1}^{N}$ 
thus contains $N$ such files in total. During training, rather than retrieving individual skills via semantic similarity, 
we select a subset $\mathcal{S} \subseteq \texttt{SkillBank}$ of $m$ skill files 
ranked by an \textit{on-policy helpfulness} criterion, which estimates the learning 
utility of each $\mathcal{S}_k$ to the current policy $\pi_\theta$ 
(detailed in Section~\ref{sec:curriculum_learning}).
at the current training stage.
\paragraph{Context Rendering.} When expanding to more domains, 
token costs become a key challenge
with both the accumulated retrieved skills and interaction history. 
Inspired by~\citet{feng2026agentocr} and \citet{shi2026memocr}, we introduce a 
\textit{context rendering} mechanism that maps the textual interaction context 
(including history $h_t$ and retrieved skills $\mathcal{S}$) to a compact RGB image. Given compression ratio $c_t$, the rendered image is encoded 
and compressed by the vision encoder \texttt{Enc} into visual representations:
\begin{equation}
    \mathcal{V}_t = \texttt{Enc}(h_t, \mathcal{S};\, c_t)
\end{equation}
where $\mathcal{V}_t \in \mathbb{R}^{d}$ serves as the compressed visual context
embedding fed into the policy, significantly reducing token overhead while
preserving the structural information necessary for decision-making.
Rather than treating the compression ratio $c_t \in (0, 1]$ as a fixed 
hyperparameter, we allow the policy to \textit{self-generate} $c_t$ at each 
step alongside the task action $a_t$:
\begin{equation}
\label{eq:rollout_icrl}
   (a_t,\, c_t) \sim \pi_{\theta}(a_t, c_t \mid I, \mathcal{V}_{t})
\end{equation}

\subsection{In-Context Reinforcement Learning (ICRL)}
\methodname{} introduces In-Context Reinforcement Learning (ICRL), which combines 
the sample efficiency and inductive bias of skill prompting with the exploration 
capability of reinforcement learning. Through a dynamic online curriculum 
(Section~\ref{sec:curriculum_learning}), skills are progressively internalized 
into the model's parameters, eliminating the need for explicit skill retrieval at inference time.

To incentivize both efficient context compression and skill internalization within 
the agent loop, we introduce a composite reward following~\citet{feng2026agentocr}, 
which jointly optimizes task success and compression efficiency. 
Let $\mathcal{I}_{\text{succ}}(\tau) \in \{0, 1\}$ denote the binary success indicator 
for trajectory $\tau$; the composite reward is defined as:
\begin{equation}
\label{eq:reward_computation}
    r^{\text{comp}}_t = 
    \begin{cases}
        \ln(c_t), & \text{if } \mathcal{I}_{\text{succ}}(\tau) = 1, \\
        0,        & \text{otherwise},
    \end{cases}
    \qquad
    \tilde{r}_t = r_t + \lambda \cdot r^{\text{comp}}_t
\end{equation}
where $c_t \in (0, 1]$ is the compression ratio at step $t$, 
and the logarithmic formulation reflects the diminishing marginal returns 
of higher compression. $r_t$ evaluates whether the agent completes the task correctly with skill enhancements at step $t$, and $\lambda \geq 0$ controls the trade-off between task performance 
and compression efficiency.

For each query $q \sim \mathcal{D}$, $\pi_{\theta_{\text{old}}}$ 
samples a group of $G$ trajectories $\{\tau_i\}_{i=1}^G$.
The training objective is:
\begin{equation}
\mathcal{L}_{\methodname}(\theta) = \;
 \mathbb{E}_{\tau_i \sim \pi_{\theta_{\text{old}}}(q),\, q \sim \mathcal{D}}
\frac{1}{\sum_{i=1}^G |\tau_i|}
\sum_{i=1}^G \sum_{t=1}^{|\tau_i|} \text{clip}(r_{i,t}(\theta),\, A_i,\, \epsilon)
- \beta \cdot \mathbb{D}_{\text{KL}}[\pi_\theta \| \pi_{\text{ref}}]
\end{equation}
where the advantage $A_i$ is computed by normalizing the total rewards 
$\{\tilde{r}(\tau_i)\}_{i=1}^G$ within the sampled group, 
and $r_{i,t}(\theta) = {\pi_\theta(\tau_{i,t} \mid q, \tau_{i,<t})}/
{\pi_{\theta_{\text{old}}}(\tau_{i,t} \mid q, \tau_{i,<t})}$ 
is the importance sampling ratio.

\subsection{Adaptive Curriculum Learning}
\label{sec:curriculum_learning}
As training progresses, the reliance on external skills undergoes a controlled \textit{annealing} process to avoid abrupt distribution shifts in the context space. We formulate this curriculum as a linear decay of the skill budget $M^{(s)}$ at each stage $s \in \{1, \dots, N_S\}$:
\begin{equation}
    |\mathcal{S}^{(s)}| \leq M^{(s)} = \left\lceil N \cdot \frac{N_S - s}{N_S - 1} \right\rceil
\end{equation}
This linear decay bounds the step-wise reduction of the skill context to $M^{(s)} - M^{(s+1)} \approx \frac{N}{N_S - 1}$. By constraining changes to the active skill set $\mathcal{S}^{(s)}$, we strictly limit the deviation in the rendered visual context $\mathcal{V}_t^{(s)} = \texttt{Enc}(h_t, \mathcal{S}^{(s)}; c_t)$. This ensures the distribution shift of the policy $\pi_\theta(a_t, c_t \mid I, \mathcal{V}_t^{(s)})$ remains smooth and stable, safely transitioning the agent to a fully self-reliant state ($\mathcal{S}^{(N_S)} = \emptyset$).
\vspace{-2mm}
\begin{figure}[h]
\centering
\begin{minipage}{0.46\columnwidth}
Based on above design, our curriculum operates in two phases: (a) an offline 
\textit{Relevance-Driven Skill Grouping} that associates each skill file 
$\mathcal{S}_k$ with a dedicated validation sub-task; and (b) an online 
\textit{Helpfulness-Driven Dynamic Curriculum} that adaptively selects the active 
skill subset $\mathcal{S}$ based on the current policy's learning state during training process.

\paragraph{(a) Relevance-Driven Skill Grouping.}
We define the \textit{relevance} between a validation sub-task and a skill file 
$\mathcal{S}_k$ as whether the sub-task's domain and objective align with the 
skill category encoded in $\mathcal{S}_k$.
Based on this relevance, we partition the validation set (subtracted from training dataset) into $N$ sub-tasks 
$\{\mathcal{T}_k\}_{k=1}^{N}$ prior to training, where $\mathcal{T}_k$ groups 
all validation instances whose skill requirements correspond to $\mathcal{S}_k$.
This offline grouping ensures each $\mathcal{S}_k$ has a dedicated sub-task 
for evaluating its utility, forming the structural basis for the subsequent 
dynamic curriculum.

\paragraph{(b) Helpfulness-Driven Dynamic Curriculum.}
We split training process into $N_S$ progressive stages 
with a decreasing skill budget $M$ ($|M|=N_S$), 
gradually reducing the agent's reliance on external skill guidance until it 
operates without any retrieved skills.
We quantify the \textit{helpfulness} metric $\Delta_k$ of each 
skill file $\mathcal{S}_k$ to the current policy $\pi_\theta$ by evaluating 
$\mathcal{T}_k$ under two conditions: with $\mathcal{S}_k$ provided 
(\textit{w/ skill}) and without it (\textit{w/o skill}) per $d$ training steps.
For stage $s$, we \texttt{Filter}, \texttt{Rank}, and \texttt{Select} top-$m$ ($m \leq M^{(s)}$) files from the active skill pool by $\Delta_k$ (see Algorithm~\ref{algorithm}).
\end{minipage}
\hfill
\begin{minipage}{0.52\columnwidth}
\refstepcounter{algorithm}
\vspace{0pt}
\hrule height 1pt
\vspace{3pt}
\small
\textbf{Algorithm \thealgorithm} \ Curriculum Learning for \methodname
\label{algorithm}
\vspace{3pt}
\hrule
\vspace{3pt}
\begin{algorithmic}[1]
\Require Initial policy $\pi_{\theta}$; reference model $\pi_{\text{ref}}$; 
         training dataset $\mathcal{D}$; 
         total training steps $T_{\text{total}}$;
         skill library $\texttt{SkillBank} = \{\mathcal{S}_k\}_{k=1}^{N}$;
         validation sub-tasks $\{\mathcal{T}_k\}_{k=1}^{N}$;
         number of stages $N_S$;
         validation interval $d$.
\Ensure Trained policy $\pi_{\theta}$

\State $M \leftarrow \left[ N,\; \left\lceil \frac{(N_S-2)}{(N_S-1)} N \right\rceil,\; \ldots,\; \left\lceil \frac{1}{N_S-1} N \right\rceil,\; 0 \right]$

\State \mycomment{Step 0: Initialize active skill subset}
\State $\mathcal{S} \leftarrow \texttt{SkillBank}$

\For{stage $s = 1, \ldots, N_S$}
    \For{step $t = 1, \ldots, \left\lfloor T_{\text{total}} / N_S \right\rfloor$}
        \If{$t \bmod d = 0$ \textbf{and} $M^{(s)} > 0$}
            \State \mycomment{Step 1: Helpfulness Evaluation for $\forall k$}
            \State $\mathrm{Acc}_k^{\text{w/ skill}} \leftarrow \texttt{Validate}(\pi_{\theta},\, \mathcal{T}_k,\, \mathcal{S})$
            \State $\mathrm{Acc}_k^{\text{w/o skill}} \leftarrow \texttt{Validate}(\pi_{\theta},\, \mathcal{T}_k,\, \emptyset)$
            \State $\Delta_k \leftarrow \mathrm{Acc}_k^{\text{w/ skill}} - \mathrm{Acc}_k^{\text{w/o skill}},$
            \State \mycomment{Step 2: Filter \& Rank}
            \State $\mathcal{S} \leftarrow \{\mathcal{S}_k \mid \Delta_k > 0\}$

            \State Sort $\mathcal{S}$ by $\Delta_k$ in descending order
            \State \mycomment{Step 3: Select top-$M^{(s)}$ skill files}
            \State $\mathcal{S} \leftarrow \mathcal{S}[1 : M^{(s)}]$
        \ElsIf{$M^{(s)} = 0$}
            \State $\mathcal{S} \leftarrow \emptyset$
        \EndIf
        \State \mycomment{Step 4: Policy update via ICRL}
        \For{$q \sim \texttt{Batched}(\mathcal{D})$}
            \State Rollout trajectories $\{\tau_i\}_{i=1}^{G}$ via Eq.~\ref{eq:rollout_icrl}
            \For{each $\tau_i$}
                \State Compute reward $\tilde{r}(\tau_i)$ via Eq.~\ref{eq:reward_computation}
                \State $A_i \leftarrow \texttt{Normalize}\bigl(\{\tilde{r}(\tau_i)\}_{i=1}^{G}\bigr)$
            \EndFor
            \State $\pi_{\theta} \leftarrow \pi_{\theta} - \nabla_{\theta}\, \mathcal{L}_{\methodname}(\theta)$
        \EndFor
    \EndFor
\EndFor
\end{algorithmic}
\vspace{3pt}
\hrule
\end{minipage}

\end{figure}

\section{Experiment}
\begin{table*}[t]
\centering
\caption{
    \textbf{Performance on ALFWorld and Search-QA tasks.} 
    We report the success rate (\%) and the average context token cost (k) per step. 
    $^\dagger$ denotes models validated with skill augmentation; 
    $^\star$ denotes methods that encodes visual context with reduced token overhead.
    We simply reproduce results of SkillRL-3B without cold start and skill evolution.
    \sethlcolor{topcolor}\hl{\textbf{Best}} and \sethlcolor{secondcolor}\hl{\mbox{\underline{second-best}}} are highlighted. 
}
\label{tab:main_results}
\resizebox{1\textwidth}{!}{%
\begin{tabular}{l CCCCCCCC CCCCCCCCC }
\toprule
& \multicolumn{8}{c}{\textbf{ALFWorld}} & \multicolumn{9}{c}{\textbf{Search-QA}} \\
\cmidrule(lr){2-9} \cmidrule(lr){10-18}
\textbf{Method}
& \textbf{Pick} & \textbf{Look} & \textbf{Clean} & \textbf{Heat} & \textbf{Cool} & \textbf{Pick2} & \textbf{Avg$\uparrow$} & \textbf{Cost$\downarrow$}
& \textbf{NQ} & \textbf{Triv} & \textbf{Pop} & \textbf{Hotp} & \textbf{2Wk} & \textbf{MuS} & \textbf{Bam} & \textbf{Avg$\uparrow$} & \textbf{Cost$\downarrow$} \\
\midrule

\rowcolor{gray!10} \multicolumn{18}{l}{\textit{Qwen2.5-(VL)-3B-Instruct}} \\

Zero-Shot
    & 27.0 & 24.3 & 4.5 & 20.5 & 10.2 & 0.0 & 15.2 & 1.21
    & 9.4 & 31.3 & 19.8 & 15.0 & 14.8 & 4.7 & 16.8 & 15.9
    & 0.48
    \\
Few-Shot$^\dagger$
    & 44.1 & 45.1 & 30.5 & 44.1 & 9.2 & 3.3 & 29.3 & 2.30
    & 11.8 & 30.9 & 20.2 & 13.7 & 18.4 & 4.5 & 25.6 & 17.9 & 0.86
    \\
Zero-Shot$^\star$
    & 44.3 & 27.6 & 8.6 & 3.1 & 5.7 & 3.1 & 17.6 & 0.48
    & 10.2 & 27.7 & 10.9 & 9.1 & 12.2 & 3.7 & 15.2 & 12.7
    & \cellcolor{topcolor}\textbf{0.15}
    \\
Few-Shot$^\star$$^\dagger$
    & 57.1 & 25.3 & 4.5 & 5.5 & 10.2 & 9.4 & 23.8 & 0.88
    & 11.1 & 26.2 & 14.2 & 15.4 & 13.4 & 3.0 & 19.2 & 14.6 & 0.36
    \\
GRPO
    & \cellcolor{secondcolor}\underline{92.6}
    & \cellcolor{secondcolor}\underline{85.7}
    & 70.6 & 86.6
    & \cellcolor{topcolor}\textbf{79.3}
    & 65.0 & 79.9 & 1.02
    & 39.3
    & \cellcolor{topcolor}\textbf{60.6}
    & 41.1
    & \cellcolor{topcolor}\textbf{37.4}
    & \cellcolor{secondcolor}\underline{34.6}
    & \cellcolor{topcolor}\textbf{15.4}
    & 26.4 & 36.4 & 0.61
    \\
AgentOCR$^\star$
    & 91.9 & 81.8 & 76.0 & 73.3 & 76.1
    & \cellcolor{secondcolor}\underline{70.0}
    & 78.2
    & \cellcolor{secondcolor}\underline{0.38}
    & 38.6 & 56.5 & 41.7 & 33.6 & 30.7
    & \cellcolor{secondcolor}\underline{14.6}
    & 24.0
    & 34.2
    & 0.26
    \\
EvolveR
    & 77.3 & 24.5 & 47.9 & 41.7 & 24.6 & 22.5 & 44.1 & 1.89
    & \cellcolor{topcolor}\textbf{43.4}
    & \cellcolor{secondcolor}\underline{58.4}
    & \cellcolor{topcolor}\textbf{43.4}
    & \cellcolor{secondcolor}\underline{37.3}
    & \cellcolor{topcolor}\textbf{38.1}
    & 13.7 & 32.8 & 38.2 & --
    \\
SkillRL$^\dagger$
    & 91.9
    & \cellcolor{topcolor}\textbf{100}
    & \cellcolor{secondcolor}\underline{82.9}
    & \cellcolor{topcolor}\textbf{87.4}
    & \cellcolor{secondcolor}\underline{78.7}
    & \cellcolor{secondcolor}\underline{70.0}
    & \cellcolor{secondcolor}\underline{82.4}
    & 2.21
    & 38.6 & 57.6 & 40.3 & 33.6 & 31.1 & 13.3
    & \cellcolor{secondcolor}\underline{58.1}
    & \cellcolor{secondcolor}\underline{38.9}
    & 0.87
    \\
\textbf{\methodname{}$^\star$}
    & \cellcolor{topcolor}\textbf{95.6}
    & 80.4
    & \cellcolor{topcolor}\textbf{100}
    & \cellcolor{secondcolor}\underline{86.7}
    & \cellcolor{secondcolor}\underline{78.7}
    & \cellcolor{topcolor}\textbf{75.2}
    & \cellcolor{topcolor}\textbf{87.9}
    & \cellcolor{topcolor}\textbf{0.38}
    & \cellcolor{secondcolor}\underline{39.8}
    & 57.5
    & \cellcolor{secondcolor}\underline{42.3}
    & 35.1 & 33.7 & 13.3
    & \cellcolor{topcolor}\textbf{63.7}
    & \cellcolor{topcolor}\textbf{40.8}
    & \cellcolor{secondcolor}\underline{0.18}
    \\

\midrule

\rowcolor{gray!10} \multicolumn{18}{l}{\textit{Qwen2.5-(VL)-7B-Instruct}} \\

Zero-Shot
    & 67.6 & 35.4 & 19.3 & 31.3 & 30.1 & 4.4 & 31.3 & 1.08
    & 10.4 & 32.4 & 22.3 & 15.8 & 15.4 & 7.2 & 19.2 & 17.5 & 0.70
    \\
Few-Shot$^\dagger$
    & 75.4 & 64.9 & 67.5 & 26.7 & 19.4 & 8.9 & 48.4 & 2.12
    & 12.3 & 36.8 & 24.5 & 17.7 & 18.2& 6.5 & 24.8& 20.1 & 0.97
    \\
Zero-Shot$^\star$
    & 46.0 & 35.6 & 19.1 & 7.1 & 5.5 & 5.4 & 21.1 & 0.52
    & 6.9 & 30.4 & 12.0 & 10.5 & 9.1 & 5.5 & 24.0 & 14.0
    & \cellcolor{topcolor}\textbf{0.26}
    \\
Few-Shot$^\star$$^\dagger$
    & 44.3 & 55.4 & 52.9 & 0.0 & 11.2 & 5.4 & 28.9 & 1.79
    & 10.5 & 31.9 & 18.7 & 14.2 & 14.4 & 6.9 & 24.8 & 17.3 & 0.41
    \\
GRPO
    & 92.6 & 93.8 & 85.2 & 80.0 & 82.7 & 56.5 & 81.8 & 0.95
    & \cellcolor{secondcolor}\underline{45.1}
    & \cellcolor{topcolor}\textbf{63.7}
    & 44.0
    & \cellcolor{topcolor}\textbf{43.6}
    & \cellcolor{topcolor}\textbf{43.2}
    & \cellcolor{secondcolor}\underline{16.8}
    & 37.6 & 41.9 & 0.73
    \\
AgentOCR$^\star$
    & \cellcolor{secondcolor}\underline{95.6}
    & \cellcolor{topcolor}\textbf{96.2}
    & 78.1 & 73.2 & 72.4 & 72.0 & 81.2
    & \cellcolor{secondcolor}\underline{0.43}
    & 43.1 & 61.0 & \cellcolor{secondcolor}\underline{45.4} & 40.8 & 38.3 & 15.7 & 36.8 & 40.1
    & 0.36
    \\
EvolveR
    & 64.9 & 33.3 & 46.4 & 13.3 & 33.3 & 33.3 & 43.8 & --
    & 43.5
    & \cellcolor{secondcolor}\underline{63.4}
    & \cellcolor{topcolor}\textbf{45.9}
    & 38.2
    & \cellcolor{secondcolor}\underline{42.0}
    & 15.6 & 54.4 & 43.1 & --
    \\
SkillRL$^\dagger$
    & 97.9
    & \cellcolor{secondcolor}\underline{71.4}
    & \cellcolor{secondcolor}\underline{90.0}
    & \cellcolor{topcolor}\textbf{90.0}
    & \cellcolor{topcolor}\textbf{95.5}
    & \cellcolor{topcolor}\textbf{87.5}
    & \cellcolor{topcolor}\textbf{89.9}
    & --
    & \cellcolor{topcolor}\textbf{45.9}
    & 63.3
    & \cellcolor{topcolor}\textbf{45.9}
    & \cellcolor{secondcolor}\underline{43.2}
    & 40.3
    & \cellcolor{topcolor}\textbf{20.2}
    & \cellcolor{topcolor}\textbf{73.8}
    & \cellcolor{topcolor}\textbf{47.1}
    & --
    \\
\textbf{\methodname{}$^\star$}
    & \cellcolor{topcolor}\textbf{100}
    & 85.8
    & \cellcolor{topcolor}\textbf{94.6}
    & \cellcolor{secondcolor}\underline{81.9}
    & \cellcolor{secondcolor}\underline{85.7}
    & \cellcolor{secondcolor}\underline{80.1}
    & \cellcolor{secondcolor}\underline{89.8}
    & \cellcolor{topcolor}\textbf{0.41}
    & 42.7 & 61.1
    & 45.3
    & 40.0 & 38.3 & 16.4
    & \cellcolor{secondcolor}\underline{66.9}
    & \cellcolor{secondcolor}\underline{44.4}
    & \cellcolor{secondcolor}\underline{0.34}
    \\
\bottomrule
\end{tabular}
}

\end{table*}

\begin{table*}[t]
\centering
\caption{
    \textbf{Performance on WebShop (128 tasks).}
    We report the Score (\%), Acc (\%) and the average context token cost (k) per step. All methods encode visual context with reduced token overhead.
}
\label{tab:webshop_results}
\resizebox{0.95\textwidth}{!}{%
\begin{tabular}{l c ccc ccc}
\toprule
& \textbf{Inference} & \multicolumn{3}{c}{\textbf{Qwen2.5-VL-3B}} & \multicolumn{3}{c}{\textbf{Qwen2.5-VL-7B}} \\
\cmidrule(lr){3-5} \cmidrule(lr){6-8}
\textbf{Method} & \textbf{w/ Skills}
& \textbf{Score$\uparrow$} & \textbf{Accuracy$\uparrow$} & \textbf{Tokens$\downarrow$}
& \textbf{Score$\uparrow$} & \textbf{Accuracy$\uparrow$} & \textbf{Tokens$\downarrow$} \\
\midrule

\rowcolor{gray!10} \multicolumn{8}{l}{\textit{Baselines}} \\

Zero-Shot
    &
    & 23.4 & 6.3  & 0.46k
    & 26.8 & 7.8  & 0.48k
    \\
Few-Shot
    & \checkmark
    & 45.8 {\small\textcolor{teal}{[+22.4]}} & 18.4 {\small\textcolor{teal}{[+12.1]}} & 0.78k {\small\textcolor{red}{[+0.52k]}}
    & 51.2 {\small\textcolor{teal}{[+24.4]}} & 24.2 {\small\textcolor{teal}{[+16.4]}} & 0.91k {\small\textcolor{red}{[+0.43k]}}
    \\
AgentOCR
    &
    & 75.2 {\small\textcolor{teal}{[+51.8]}} & 56.3 {\small\textcolor{teal}{[+50.0]}} & 0.42k {\small\textcolor{teal}{[-0.04k]}}
    & 78.6 {\small\textcolor{teal}{[+51.8]}} & 59.3 {\small\textcolor{teal}{[+51.5]}} & 0.40k {\small\textcolor{teal}{[-0.08k]}}
    \\

\midrule

\rowcolor{gray!10} \multicolumn{8}{l}{\textit{Ours: RL w/ Skills}} \\

\textbf{\methodname{}}
    & \checkmark
    & 80.9 {\small\textcolor{teal}{[+57.5]}} & 64.1 {\small\textcolor{teal}{[+57.8]}} & 0.94k {\small\textcolor{red}{[+0.48k]}}
    & 85.3 {\small\textcolor{teal}{[+58.5]}} & 71.9 {\small\textcolor{teal}{[+64.1]}} & 0.89k {\small\textcolor{red}{[+0.41k]}}
    \\
\textbf{\methodname{}}
    &
    & 78.6 {\small\textcolor{teal}{[+55.2]}} & 66.4 {\small\textcolor{teal}{[+60.1]}} & 0.49k {\small\textcolor{red}{[+0.03k]}}
    & 85.1 {\small\textcolor{teal}{[+58.3]}} & 74.2 {\small\textcolor{teal}{[+66.4]}} & 0.46k {\small\textcolor{teal}{[-0.02k]}}
    \\

\bottomrule
\end{tabular}
}
\end{table*}

\subsection{Experiment Setup}
\paragraph{Benchmarks}

We evaluate our methods on ALFWorld~\citep{shridhar2020alfworld}, Search-based QA~\citep{jin2025searchr1}, and Webshop~\citep{yao2022webshop}. \textit{ALFWorld} is a text-based game aligned with the ALFRED embodied AI benchmark, including 3,827 task instances across six categories of common household activities: Pick and Place (Pick), Look at Obj in Light (Look), Pick Clean then Place in Recep (Clean), Pick Heat then Place in Recep (Heat), Pick Cool then Place in Recep (Cool), and Pick Two Obj and Place (Pick2). \textit{Search-based QA} contains several widely-used search-augmented QA benchmarks, including single-hop QA datasets (NQ~\citep{kwiatkowski2019nq}, TriviaQA~\citep{joshi2017triviaqa}, and PopQA~\citep{mallen2023popqa}) and multi-hop QA datasets (HotpotQA~\citep{yang2018hotpotqa}, 2Wiki~\citep{ho20202wiki}, MuSiQue~\citep{trivedi2022musique}, and Bamboogle~\citep{press2023bamboogle}).
WebShop is a complex, web-based interactive environment designed to test the LLM agents in realistic online shopping scenarios. Agents navigate a realistic web interface to find and purchase products matching user specifications. We select 128 fixed tasks in validation set, which aligns with \citet{feng2025gigpo}.
\paragraph{Baselines.}
We first compare \methodname with in-context skills prompting methods (with text and OCR-based history) and RL-based methods (GRPO~\citep{shao2024deepseekmath}, AgentOCR~\citep{feng2026agentocr}, EvolveR~\citep{wu2025evolver}, and SkillRL~\citep{xia2026skillrl}) across both benchmarks in Table~\ref{tab:main_results}. For ALFWorld only (as shown in Table~\ref{tab:alfworld}), we additionally report prompt-based agentic or memory-based methods, including ReAct~\citep{yao2022react} and Reflexion~\citep{shinn2024reflexion}, as well as Mem0~\citep{chhikara2025mem0}, ExpeL~\citep{zhao2024expel}, MemP~\citep{fang2025memp}, MemRL~\citep{zhang2026memrl}, and SimpleMem~\citep{liu2026simplemem}. For search-augmented QA (as shown in Table~\ref{tab:searchqa}), we include Search-o1~\citep{li2025searcho1}, Search-R1~\citep{jin2025searchr1}, ZeroSearch~\citep{sun2025zerosearch}, O$^2$-Searcher~\citep{mei20252o2searcher}, ParallelSearch~\citep{zhao2025parallelsearch} and StepSearch~\citep{wang2025stepsearch}. Some closed-source models are also included, such as GPT-4o~\citep{hurst2024gpt-4o} and Gemini-2.5-Pro~\citep{comanici2025gemini}.

\paragraph{Implementation Details.}
We train the Qwen2.5-VL series using \methodname{} for at most 180 steps on 4 H800 GPUs.
For ALFWorld, we adopt the training data split from GiGPO~\citep{feng2025gigpo}, 
with each batch sampling 16 tasks and 8 rollouts per prompt, 
and a maximum prompt length of 3,072 tokens.
For Search-QA, we follow the experimental setup of Search-R1~\citep{jin2025searchr1}, 
using E5~\citep{wang2022e5} as the retriever.
The training data are drawn from NQ and HotpotQA, making these two benchmarks in-domain, 
while the remaining datasets serve as out-of-domain evaluation.
Each batch samples 128 tasks with a maximum prompt length of 4,096 tokens.
For Webshop, 1000 tasks are selected for training, with each batch sampling 16 tasks and 8 rollouts per prompt, and a maximum prompt length of 4,096 tokens. 
For the curriculum learning schedule, we set the validation subset size to 1,000, 
the number of curriculum stages to $N_S = 3$, 
and initialize \texttt{SkillBank} from SkillRL~\citep{xia2026skillrl} for both environments.

\subsection{Main Results}

\paragraph{Method Performance.}
As shown in Table~\ref{tab:main_results} and Table~\ref{tab:webshop_results},
\methodname{} demonstrates exceptional performance across ALFWorld, Search-QA, and
WebShop.
While introducing explicit skill prompts (Few-Shot) brings moderate improvements over
Zero-Shot baselines, the gains are limited, indicating that LLMs struggle to fully
leverage skill descriptions without sufficient exploration.
In contrast, without external skill prompting during inference, \methodname{} (3B)
achieves an average success rate of 87.9 on ALFWorld, 40.8 on Search-QA, and a score
of 78.6 with 66.4\% accuracy on WebShop, outperforming AgentOCR by +9.7, +6.6, and
+10.1 (accuracy) respectively.
Based on 7B models, it delivers scores of 89.8 on ALFWorld, 44.4 on Search-QA, and
85.1/74.2 (score/accuracy) on WebShop, substantially outperforming other RL-based
methods such as EvolveR, AgentOCR, and GRPO.
Notably, on WebShop the skill-free \methodname{} (7B) surpasses AgentOCR by +6.5 in
score and +14.9 in accuracy, demonstrating strong internalization of complex
e-commerce navigation behaviors.
Furthermore, \methodname{} achieves competitive or even stronger performance against
skill-augmented methods like SkillRL.
These consistent gains over both zero-shot and skill-augmented baselines across three
diverse benchmarks demonstrate that our approach successfully \textit{internalizes}
complex reasoning and tool-use behaviors into the model's parameters.

Table~\ref{tab:alfworld} and Table~\ref{tab:searchqa} provide broader comparisons.
On ALFWorld, \methodname{} (89.8) largely outperforms memory-augmented learning
methods, including ExpeL (46.3), Mem0 (54.7), and MemRL (21.4).
On Search-based QA, \methodname{} (44.4) likewise surpasses search-based methods such
as Search-R1 (38.5), ZeroSearch (39.1), and EvolveR (43.1), further highlighting its
generality.

\paragraph{Token Efficiency.}
Beyond strong task performance, \methodname{} achieves these results with a
substantially lower context token cost.
Due to visual context modeling and skill internalization, \methodname{} fundamentally
maintains an ultra-low average token cost per step.
For instance, using 3B models, it consumes only 0.38k tokens per step on ALFWorld,
0.18k on Search-QA, and 0.49k on WebShop.
This is a massive reduction compared to text-based or skill-augmented methods like
SkillRL, which costs 2.21k and 0.87k tokens per step on ALFWorld and Search-QA
respectively (more than \textbf{5$\times$} higher).
On WebShop, skill-augmented Few-Shot inference requires 0.78--0.91k tokens per step,
while \methodname{} without skills achieves superior accuracy at only 0.46--0.49k
tokens---a nearly \textbf{2$\times$} reduction in context cost while delivering over
\textbf{3$\times$} the accuracy of the Few-Shot baseline.

\subsection{Training Dynamics}
\begin{figure}[t]
\centering
\begin{minipage}{0.48\columnwidth}
    \centering
    \includegraphics[width=\linewidth]{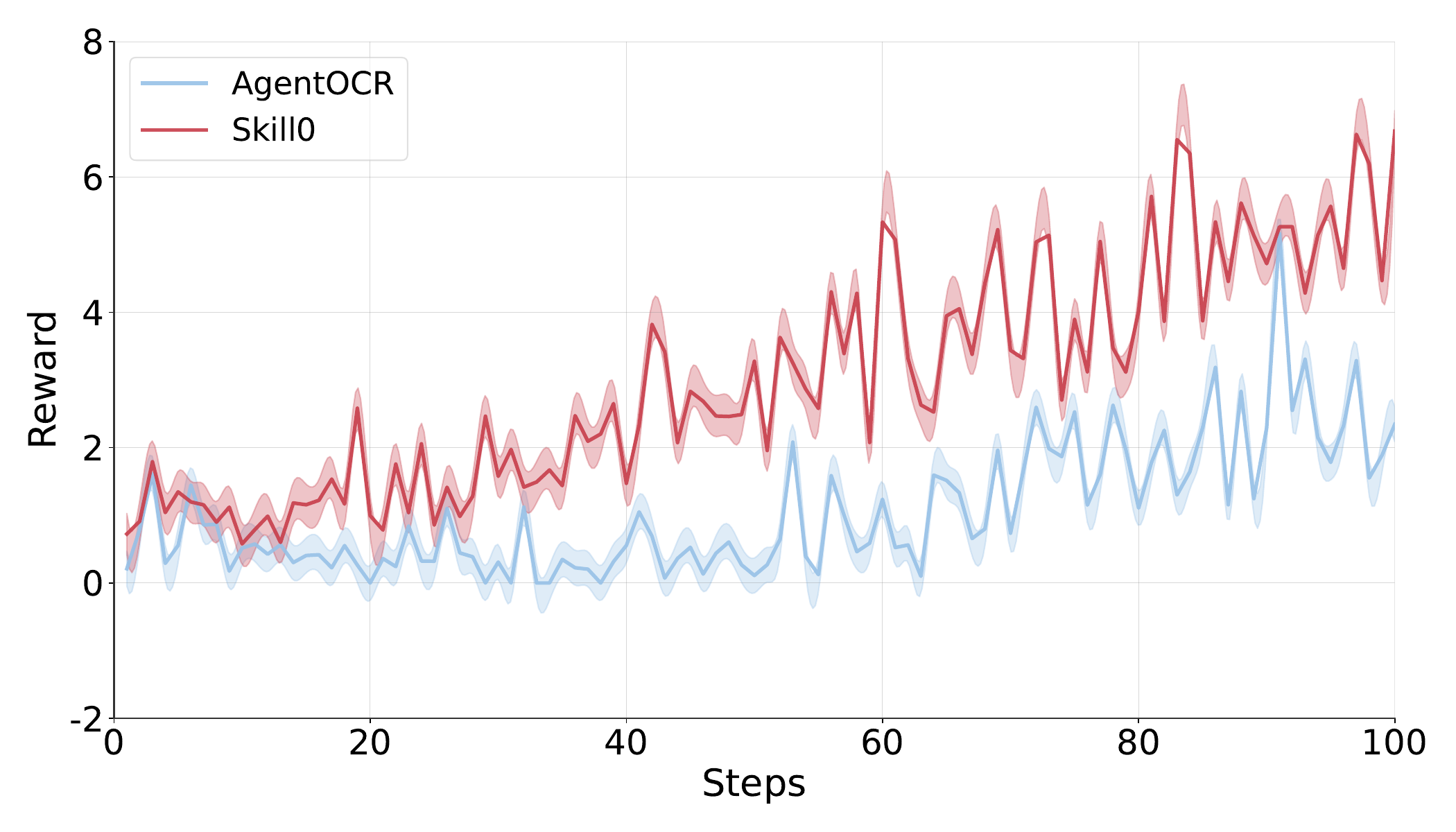}
    \caption{Comparison of training dynamics with AgentOCR on Qwen2.5-VL-3B.}
    \label{fig:training_reward_3b}
\end{minipage}
\hfill
\begin{minipage}{0.48\columnwidth}
    \centering
    \includegraphics[width=\linewidth]{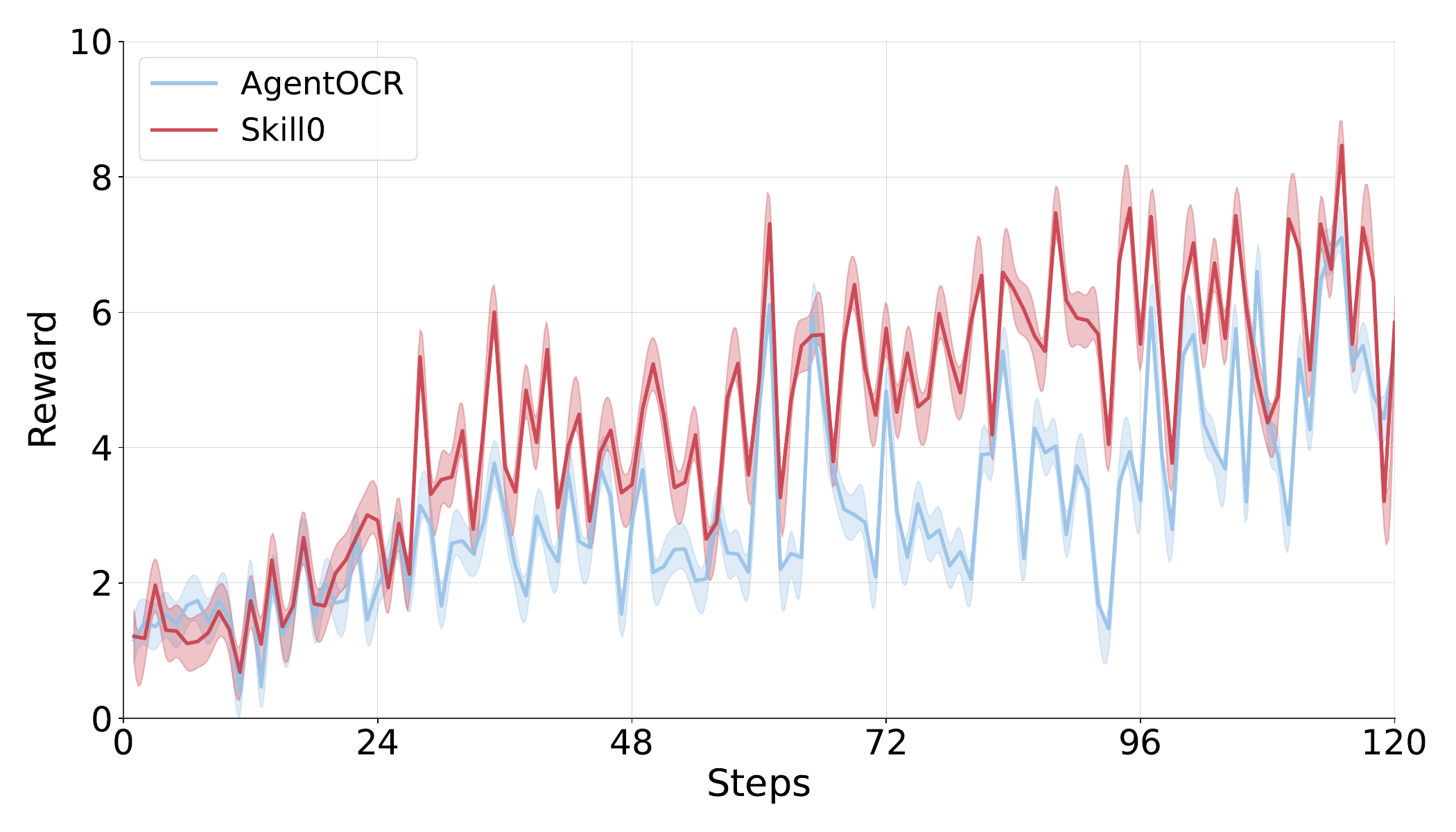}
    \caption{Comparison of training dynamics with AgentOCR on Qwen2.5-VL-7B.}
    \label{fig:training_reward_7b}
\end{minipage}
\end{figure}

\begin{figure}[t]
\centering
\includegraphics[width=\columnwidth]{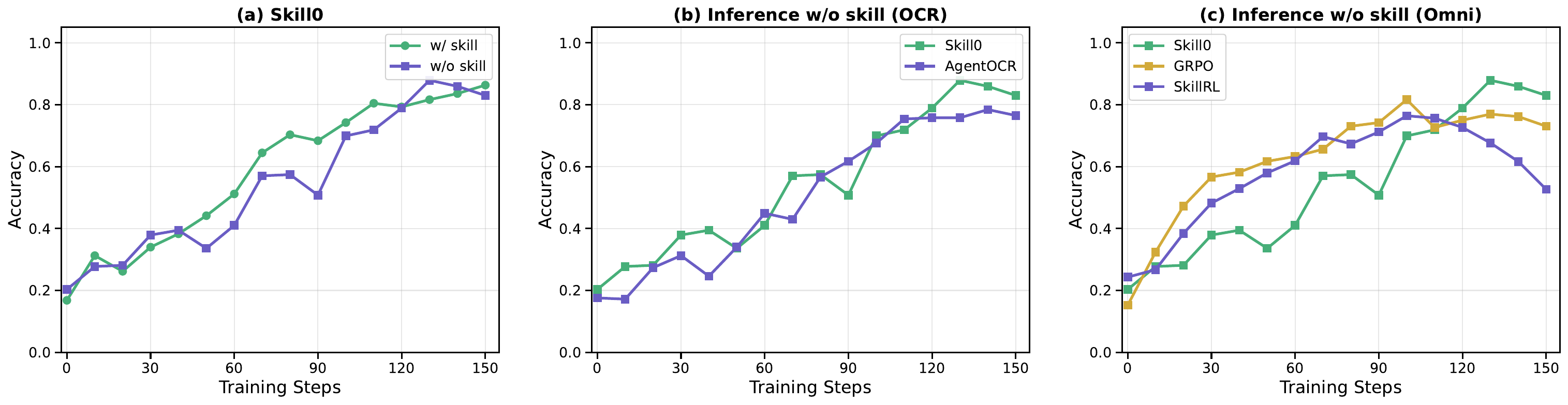}
\caption{\textbf{Training Dynamics Comparison.} (a) Validation performance of \methodname{} (OCR) with and without skill augmentation, evaluated every 10 training steps. (b) Performance comparison between \methodname{} and AgentOCR, both evaluated without skill augmentation. (c) Performance comparison of \methodname{} (OCR) against GRPO (Text) and SkillRL (Text), all evaluated without skill augmentation.}
\label{fig:skill_comparison_1x3}
\end{figure}
\paragraph{Reward.} Throughout RL optimization, \methodname{} maintains consistently higher reward curves on both the 3B and 7B backbones compared to the AgentOCR baseline, as illustrated in Figure~\ref{fig:training_reward_3b} and~\ref{fig:training_reward_7b}.

\paragraph{Method Comparison.} We further monitor validation accuracy over the course of training in Figure~\ref{fig:skill_comparison_1x3}. \textit{(a)} demonstrates that when validated \emph{with} skill augmentation, the model achieves faster early-stage performance improvement; while validation \emph{without} skill prompts yields lower initial performance, it gradually catches up toward the end of optimization, revealing a clear trend of \textit{skill internalization}.  To further validate this observation, \textit{(b)} evaluates models \emph{without} skill prompts at inference time under a strictly fair comparison setting: \methodname{} still outperforms AgentOCR, confirming that the performance advantage stems from internalized knowledge rather than reliance on explicit skill descriptions. For a broader comparison, \textit{(c)} contrasts \methodname{} against standard text-based RL baselines under the same skill-free inference protocol. Unlike GRPO and SkillRL, which plateau relatively early in training, \methodname{} continues to improve steadily throughout optimization, ultimately achieving the highest performance upper bound among all compared methods. We also provide subtask dynamics of \methodname in Appendix~\ref{sec:more_training_dynamics} to further support it.

\begin{figure}[t]
\centering
\includegraphics[width=\columnwidth]{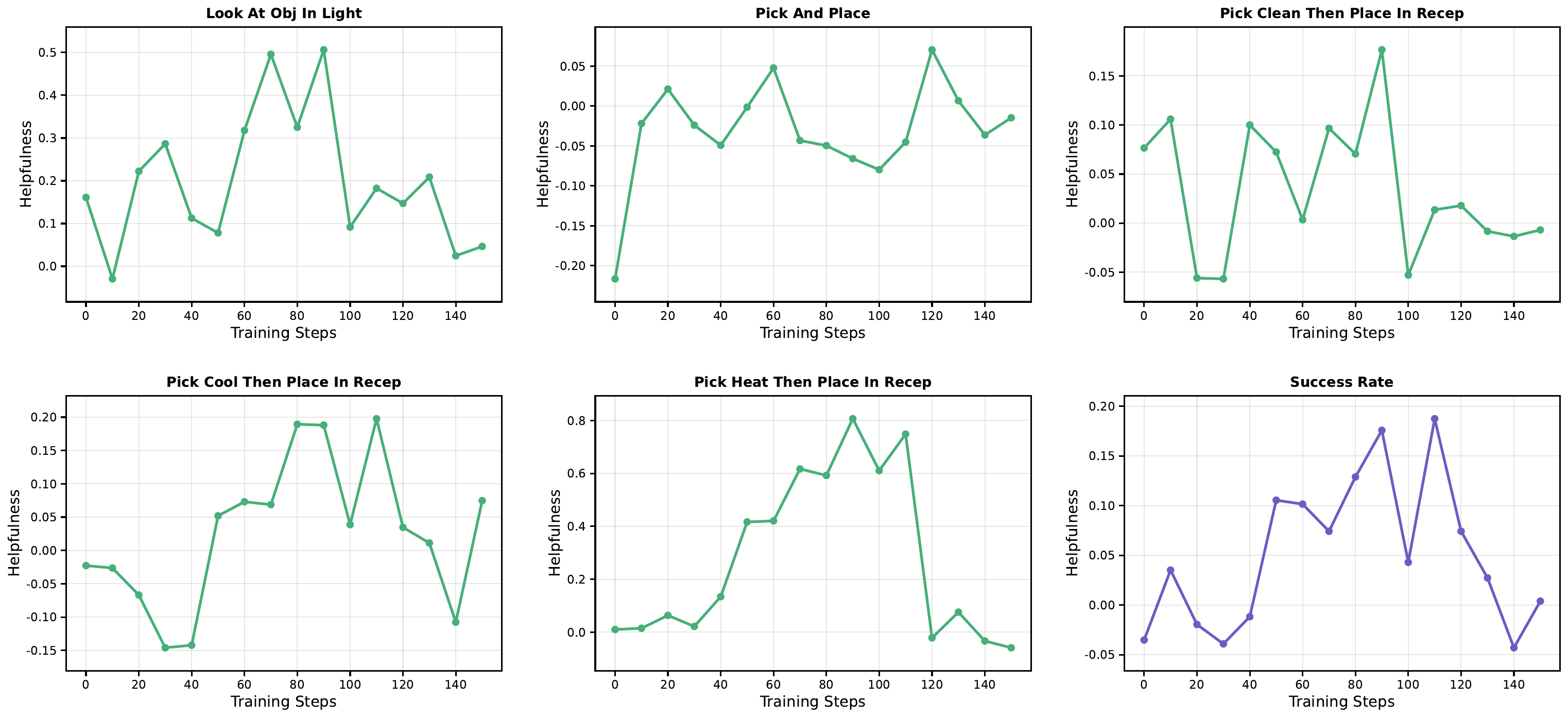}
\caption{\textbf{Training Dynamics of \textit{Helpfulness},} which are reported by $\Delta_k$ for each sub-task $k$.}
\label{fig:helpfulness_dynamics}
\end{figure}

\paragraph{Helpfulness.}
Figure~\ref{fig:helpfulness_dynamics} illustrates that the \textit{helpfulness} 
of skills exhibits a consistent \emph{rise-then-fall} pattern across all sub-tasks 
throughout training. In the early stages, \textit{helpfulness} remains low, as the 
policy has not yet learned to leverage skill prompts via direct in-context prompting. As training progresses, the policy gradually learns to ground its 
actions in the provided skill context, leading to a steady increase in 
\textit{helpfulness}. In the later stages, the dynamic curriculum progressively 
reduces the skill budget, compelling the policy to internalize skill knowledge 
into its parameters rather than relying on external prompts; consequently, 
$\Delta_k$ converges back toward zero. This characteristic trajectory empirically 
validates the synergistic working mechanism of ICRL and curriculum learning, 
demonstrating that skills serve as effective yet transient scaffolding during 
policy optimization.

\begin{figure}[t]
\centering
\begin{minipage}{0.4\columnwidth}
    \centering
    \includegraphics[width=\linewidth]{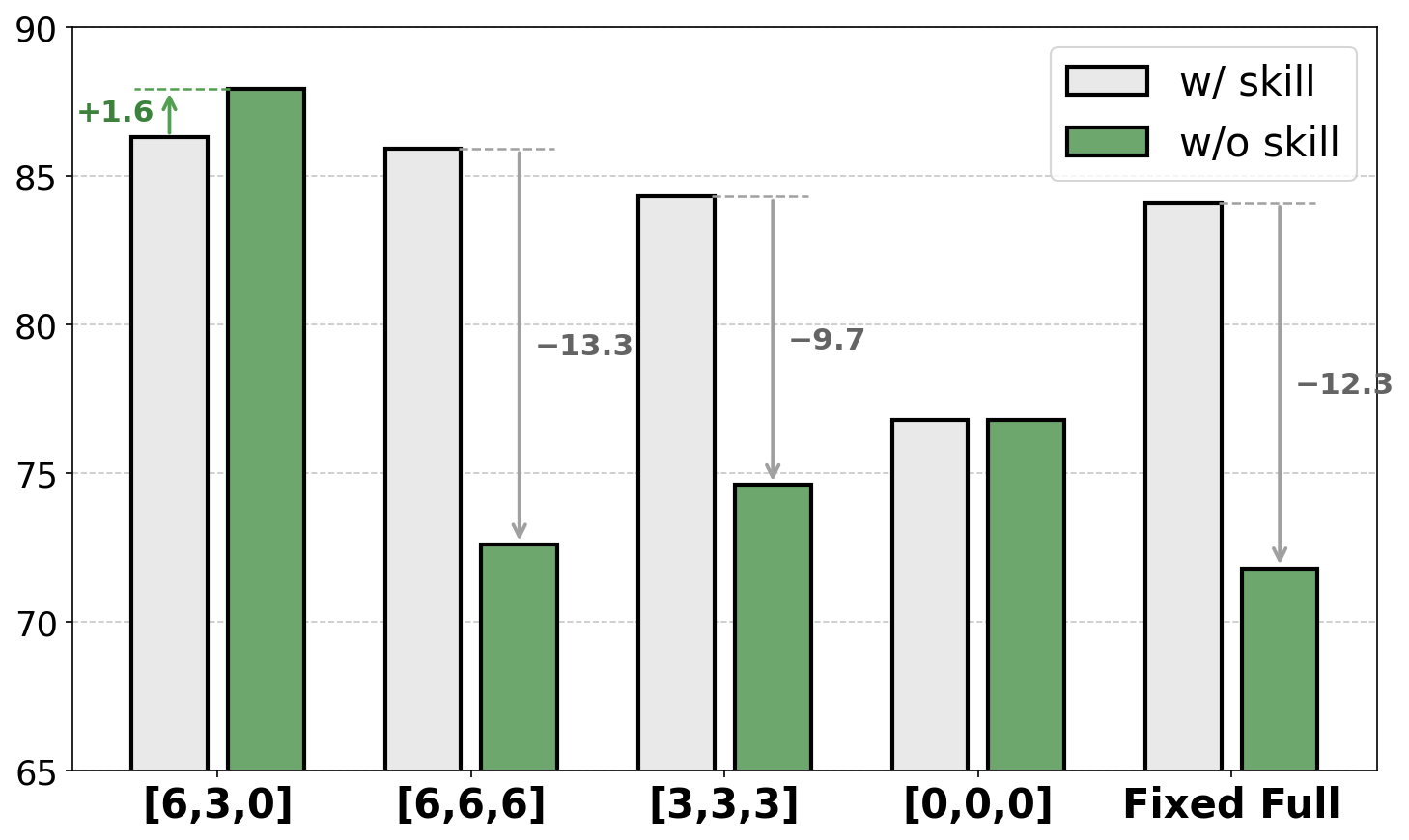}
    \caption{Ablations of skill budget $M$.}
    \label{fig:ablations_skill_budget}
\end{minipage}
\hfill
\begin{minipage}{0.58\columnwidth}
    \centering
    \includegraphics[width=\linewidth]{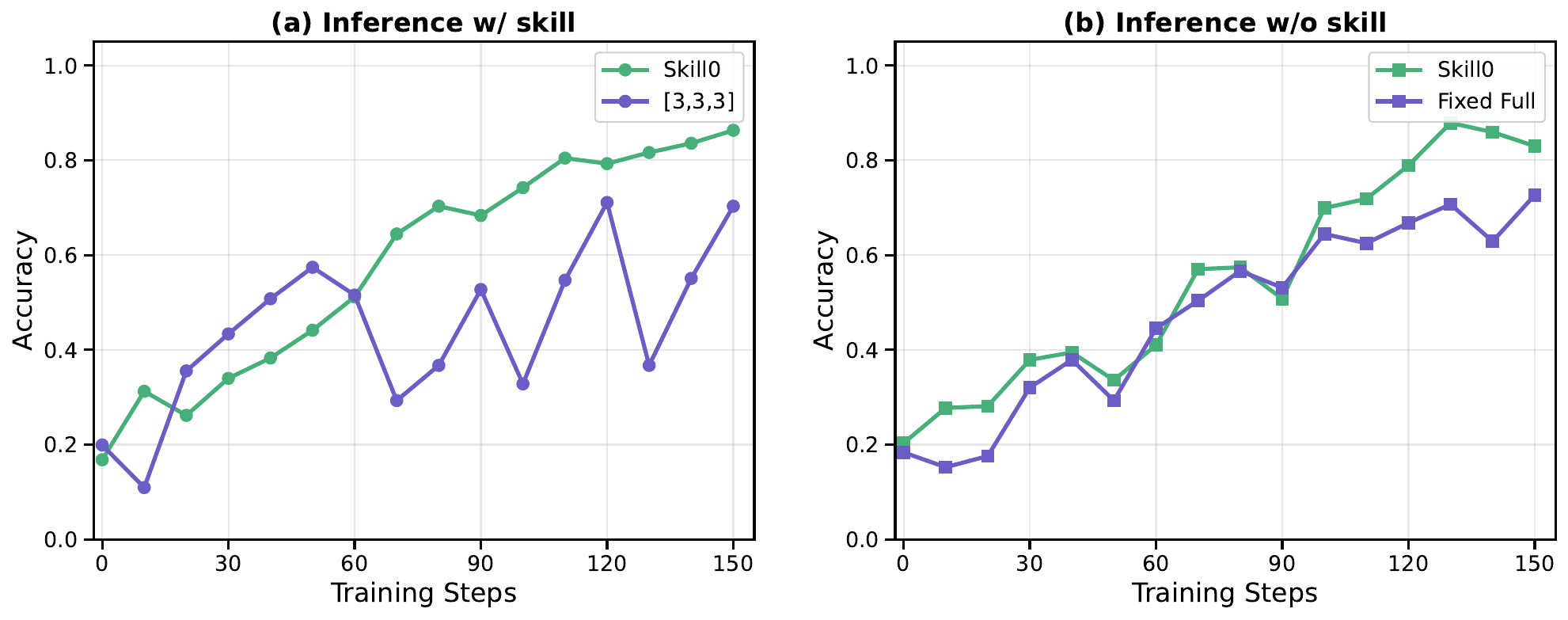}
    \caption{Ablations of skill budget during training process.}
    \label{fig:ablations_skill_budget_training}
\end{minipage}
\end{figure}

\subsection{Ablations}
\paragraph{Skill Budget $M$.} Given $N_S$ as 3, the Skill Budget $M$ for ALFWorld is calculated as $[6,3,0]$ (and $[5,3,0]$ for Search-QA) according to Algorithm~\ref{algorithm}.
This design enforces a hard upper bound that gradually anneals the number of available skills across curriculum stages, compelling the model to actively and flexibly prune less helpful skills within the Budget, thereby progressively internalizing skill knowledge. To validate the effectiveness of this design, we compare our $[6,3,0]$ against other budgets ($[6,6,6], [3,3,3]$, and $[0,0,0]$) as well as a Fixed Full setting (without filter). Figure~\ref{fig:ablations_skill_budget} highlights our superior skill internalization: while Fixed Full and $[6,6,6]$ collapse by -12.3 and -13.3 when skill prompts are removed, our method even achieves a +1.6 gain. Training dynamics in Figure~\ref{fig:ablations_skill_budget_training}~(a) show that a static low budget ($[3,3,3]$) limits early exploration, leading to unstable learning and lower peaks. Conversely, Figure~\ref{fig:ablations_skill_budget_training}~(b) demonstrates that our curriculum strategy consistently outperforms Fixed Full in skill-free inference settings, likely due to the training-inference gap and the skill over-reliance induced by maintaining a constant full skill set throughout training.
\paragraph{Dynamic Curriculum.}
Table~\ref{tab:ablation_method} validates the necessity of our three-step helpfulness-driven strategy (\texttt{Filter} \& \texttt{Rank} \& \texttt{Select}) (Algorithm~\ref{algorithm}, \textit{Step 1-3}). It achieves the highest performance (87.9\% w/o $\mathcal{S}$) and is the only setting to show a positive transfer ($\Delta = +1.6\%$) when skill prompts are removed at inference. In contrast, simply using all skills up to the budget (``w/o \texttt{Filter}'') introduces context noise that drops performance by $2.7\%$. Worse, selecting skills randomly (``w/o \texttt{Rank}'') causes a severe collapse ($\Delta = -13.7\%$, dropping to 62.9\%), proving that retaining strictly \textit{helpful} skills is essential for stable policy learning and preventing superficial prompt dependency.
\paragraph{Validation Interval $d$.}
Table~\ref{tab:ablation_internal} explores the impact of the validation interval $d$ used for helpfulness evaluation. While a smaller interval ($d=5$) provides marginal gains on Search-QA, it incurs significantly higher computational overhead.  We select $d=10$ as the optimal trade-off, balancing high task performance with training efficiency.

We also provide more detailed ablation results in Table~\ref{tab:detailed_ablations} to demonstrate our careful design.

\begin{table}[t]
    \centering

    \begin{minipage}{0.58\textwidth}
        \centering
            \caption{Ablations of Dynamic Skill Curriculum on ALFWorld in different inference settings.}
        \label{tab:ablation_method}
        \begin{tabular}{lccc}
            \toprule
            Method & w/ $\mathcal{S}$ & w/o $\mathcal{S}$ & $\Delta$ \\
            \midrule
            \rowcolor{gray!10} \texttt{Filter} \& \texttt{Rank} \& \texttt{Select} & 86.3 & 87.9 & \posval{$\uparrow$1.6} \\
            \quad w/o \texttt{Filter} & 81.6 & 78.9 & \negval{$\downarrow$2.7} \\
            \quad w/o \texttt{Rank} (Random Select)  & 76.6 & 62.9 & \negval{$\downarrow$13.7}\\
            \bottomrule
        \end{tabular}

    \end{minipage}
    \hfill 
    \begin{minipage}{0.4\textwidth}
        \centering
                \caption{Impact of Validation Interval $d$ on ALFWorld and Search-QA (subset).}
        \label{tab:ablation_internal}
        \begin{tabular}{lcc}
            \toprule
            $d$ & ALFWorld & Search-QA \\
            \midrule
            \rowcolor{gray!10} 10 & 87.9 & 48.9 \\
            5 & 87.5 & 49.6 \\
            20 & 78.1 & 42.3 \\
            \bottomrule
        \end{tabular}

    \end{minipage}
\end{table}
\section{Conclusion}
In this work, we proposed \textbf{\methodname{}}, an in-context reinforcement learning 
framework that internalizes agent skills directly into model parameters via a 
\textbf{Dynamic Curriculum} mechanism, eliminating external skill reliance at inference time.
Extensive experiments across ALFWorld and Search-QA demonstrate substantial improvements 
over RL baselines (+9.7, +6.6, and +10.1 respectively) with fewer than 0.5k tokens per step, 
establishing skill internalization as a principled alternative to the inference-time skill augmentation paradigm. We believe \methodname{} establishes skill internalization as a new principled and 
scalable paradigm, 
paving the way from tool-augmented toward truly autonomous LLM agents and self-sufficient intelligence.
\paragraph{Limitations.} \methodname{} relies
on the quality of the initial \texttt{SkillBank}, and the offline relevance-driven 
skill grouping requires re-partitioning when applied to new task domains.
\bibliographystyle{main}   
\newpage
\bibliography{reference}
\newpage
\appendix

\newpage
\section{Theoretical Analysis}
\label{sec:theoretical_analysis}

Given Eq.~\ref{eq:sufficient_stages}, each stage $s \in \{1, \dots, N_S\}$ satisfies:
\begin{equation}
    |\mathcal{S}^{(s)}| \leq M^{(s)} = \left\lceil N \cdot \frac{N_S - s}{N_S - 1} \right\rceil
\end{equation}

\subsection{Stability Analysis}

The advantage for trajectory $\tau_i$ is estimated via group-level normalization:
\begin{equation}
\label{eq:grpo_advantage}
    A_i = \frac{\tilde{r}(\tau_i) - \mu_G}{\sigma_G}, \quad
    \mu_G = \frac{1}{G}\sum_{j=1}^G \tilde{r}(\tau_j), \quad
    \sigma_G = \sqrt{\frac{1}{G}\sum_{j=1}^G (\tilde{r}(\tau_j) - \mu_G)^2}
\end{equation}
An abrupt skill context shift between stages would degrade the signal-to-noise ratio
of $A_i$. We now quantify how the linear budget schedule controls this effect.

Under the linear budget schedule, the number of skills removed at each stage transition
is uniformly bounded:
\begin{equation}
\label{eq:removal_bound}
    n_{\Delta}^{(s)} := |\mathcal{S}^{(s)} \setminus \mathcal{S}^{(s+1)}|
    \leq M^{(s)} - M^{(s+1)}
    \leq \left\lceil \frac{N}{N_S - 1} \right\rceil
\end{equation}
This follows from $M^{(s)} - M^{(s+1)} = \lceil N(N_S-s)/(N_S-1)\rceil -
\lceil N(N_S-s-1)/(N_S-1)\rceil \leq \lceil N/(N_S-1)\rceil$,
using $\lceil a \rceil - \lceil b \rceil \leq \lceil a-b \rceil$ for $a \geq b \geq 0$.

Let $R^{(s)}(q) := \tilde{r}(\tau)$ denote the reward under context
$\mathcal{V}_t^{(s)}$ for query $q$. Assuming each skill file $\mathcal{S}_k$ has
bounded influence on the expected reward:
\begin{equation}
\label{eq:reward_shift}
    \sup_{q \in \mathcal{D}} \left|
    \mathbb{E}[R^{(s+1)}(q)] - \mathbb{E}[R^{(s)}(q)]
    \right|
    \leq n_{\Delta}^{(s)} \cdot \delta_r
    \leq \left\lceil \frac{N}{N_S - 1} \right\rceil \delta_r
\end{equation}

Following~\citet{shao2024deepseekmath}, the per-step policy improvement satisfies:
\begin{equation}
\label{eq:grpo_convergence}
    J(\theta_{s+1}) - J(\theta_s) \geq \frac{1}{1-\gamma}\left(
    \mathbb{E}_{a \sim \pi_{\theta_{s+1}}}[A^{\pi_{\theta_s}}(a)]
    - \frac{2\epsilon\gamma}{1-\gamma} \cdot D_{\mathrm{TV}}^{\max}(\pi_{\theta_{s+1}}, \pi_{\theta_s})
    \right)
\end{equation}
where $\epsilon = \max_{a}|A^{\pi_{\theta_s}}(a)|$. A stage transition introduces
non-stationarity of magnitude $\lceil N/(N_S-1)\rceil\delta_r$ (Eq.~\ref{eq:reward_shift}),
yielding advantage bias:
\begin{equation}
\label{eq:advantage_bias}
    |\text{Bias}(A_i^{(s \to s+1)})| \leq \frac{n_{\Delta}^{(s)} \cdot \delta_r}{\sigma_G^{(s)}}
\end{equation}
The estimator remains reliable when (i) $n_{\Delta}^{(s)}$ is small (controlled by
$N_S$), or (ii) $\sigma_G^{(s)}$ is large relative to the perturbation.
By~\citet{guo2025ds-r1}, GRPO retains monotonic improvement when:
\begin{equation}
\label{eq:stability_condition}
    \frac{\lceil N/(N_S - 1) \rceil \cdot \delta_r}{\sigma_G} < \epsilon_{\text{clip}}
\end{equation}
This yields a sufficient condition on the number of stages:
\begin{equation}
\label{eq:sufficient_stages}
    N_S > 1 + \frac{N \cdot \delta_r}{\epsilon_{\text{clip}} \cdot \sigma_G}
\end{equation}

\subsection{Skill Selection Optimality}

For skill file $\mathcal{S}_k$ and active set $\mathcal{S}$, the marginal helpfulness
is defined as:
\begin{equation}
\label{eq:marginal_help}
    \Delta_k(\mathcal{S}) :=
    \mathrm{Acc}(\pi_\theta,\, \mathcal{T}_k,\, \mathcal{S})
    - \mathrm{Acc}(\pi_\theta,\, \mathcal{T}_k,\, \mathcal{S} \setminus \{\mathcal{S}_k\})
\end{equation}
The selection objective at stage $s$ is:
\begin{equation}
\label{eq:selection_obj}
    \mathcal{S}^{*(s)} = \argmax_{\mathcal{S} \subseteq \texttt{SkillBank},\;
    |\mathcal{S}| \leq M^{(s)}} J(\mathcal{S};\, \pi_\theta)
\end{equation}
where $J(\mathcal{S};\pi_\theta) = \sum_{k:\,\mathcal{S}_k \in \mathcal{S}}
\mathrm{Acc}(\pi_\theta, \mathcal{T}_k, \mathcal{S})$.

Assuming $J(\cdot;\pi_\theta)$ satisfies $\alpha$-approximate
submodularity~\citep{horel2024maximization,nemhauser1978analysis}, i.e., for all
$\mathcal{A} \subseteq \mathcal{B} \subseteq \texttt{SkillBank}$ and
$\mathcal{S}_k \notin \mathcal{B}$:
\begin{equation}
\label{eq:submod}
    J(\mathcal{A} \cup \{\mathcal{S}_k\}) - J(\mathcal{A})
    \geq \alpha \cdot \left(J(\mathcal{B} \cup \{\mathcal{S}_k\}) - J(\mathcal{B})\right),
    \quad \alpha \in (0,1]
\end{equation}
the greedy selection of skills with $\Delta_k > 0$ in decreasing order up to budget
$M^{(s)}$ satisfies:
\begin{equation}
\label{eq:greedy_bound}
    J(\mathcal{S}^{\text{greedy}};\,\pi_\theta)
    \geq \left(1 - e^{-\alpha}\right) \cdot J(\mathcal{S}^{*(s)};\,\pi_\theta)
\end{equation}
When $\alpha=1$, Eq.~\ref{eq:greedy_bound} recovers the classical $(1-1/e)$-approximation
ratio.

As the policy internalizes skills via GRPO, helpfulness scores evolve as:
\begin{equation}
    \frac{\partial \Delta_k}{\partial \theta}
    = \frac{\partial}{\partial\theta}\left[
    \mathrm{Acc}(\pi_\theta, \mathcal{T}_k, \mathcal{S})
    - \mathrm{Acc}(\pi_\theta, \mathcal{T}_k, \emptyset)
    \right]
\end{equation}
Successful internalization of $\mathcal{S}_k$ drives
$\mathrm{Acc}(\pi_\theta, \mathcal{T}_k, \emptyset)$
to $\mathrm{Acc}(\pi_\theta, \mathcal{T}_k, \mathcal{S})$,
hence $\Delta_k$ to 0. Such skills are automatically excluded by the positivity filter
in Eq.~\ref{eq:selection_obj}, yielding a \textit{self-paced} curriculum where $M^{(s)}$
serves as an upper bound.
\section{More Training Dynamics}
\label{sec:more_training_dynamics}
\begin{figure}[h]
\centering
\includegraphics[width=\columnwidth]{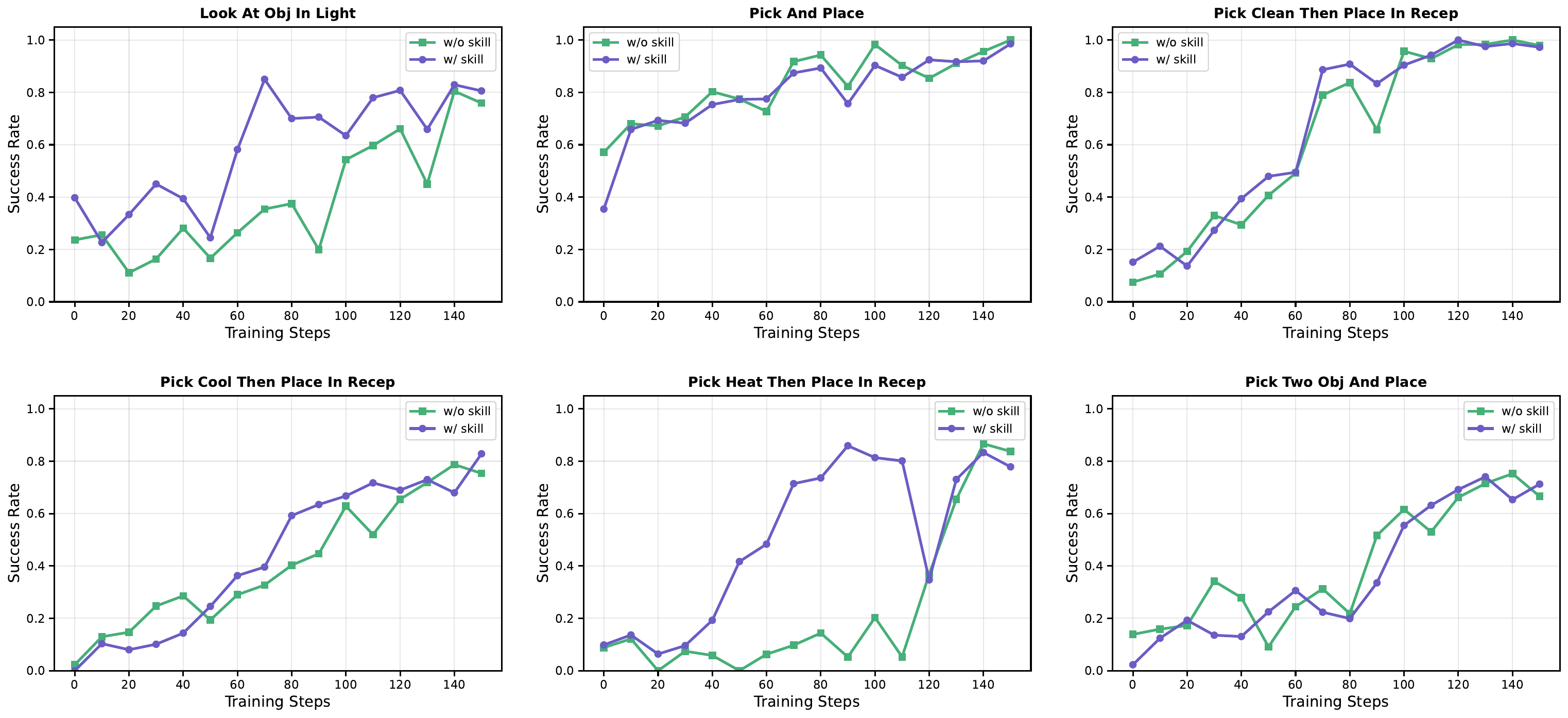}
\caption{Training dynamics of \methodname on Qwen2.5VL-3B, with ALFWorld accuracy reported.}
\label{fig:subtask_comparison_alfworld}
\end{figure}

\begin{figure}[h]
\centering
\includegraphics[width=0.8\columnwidth]{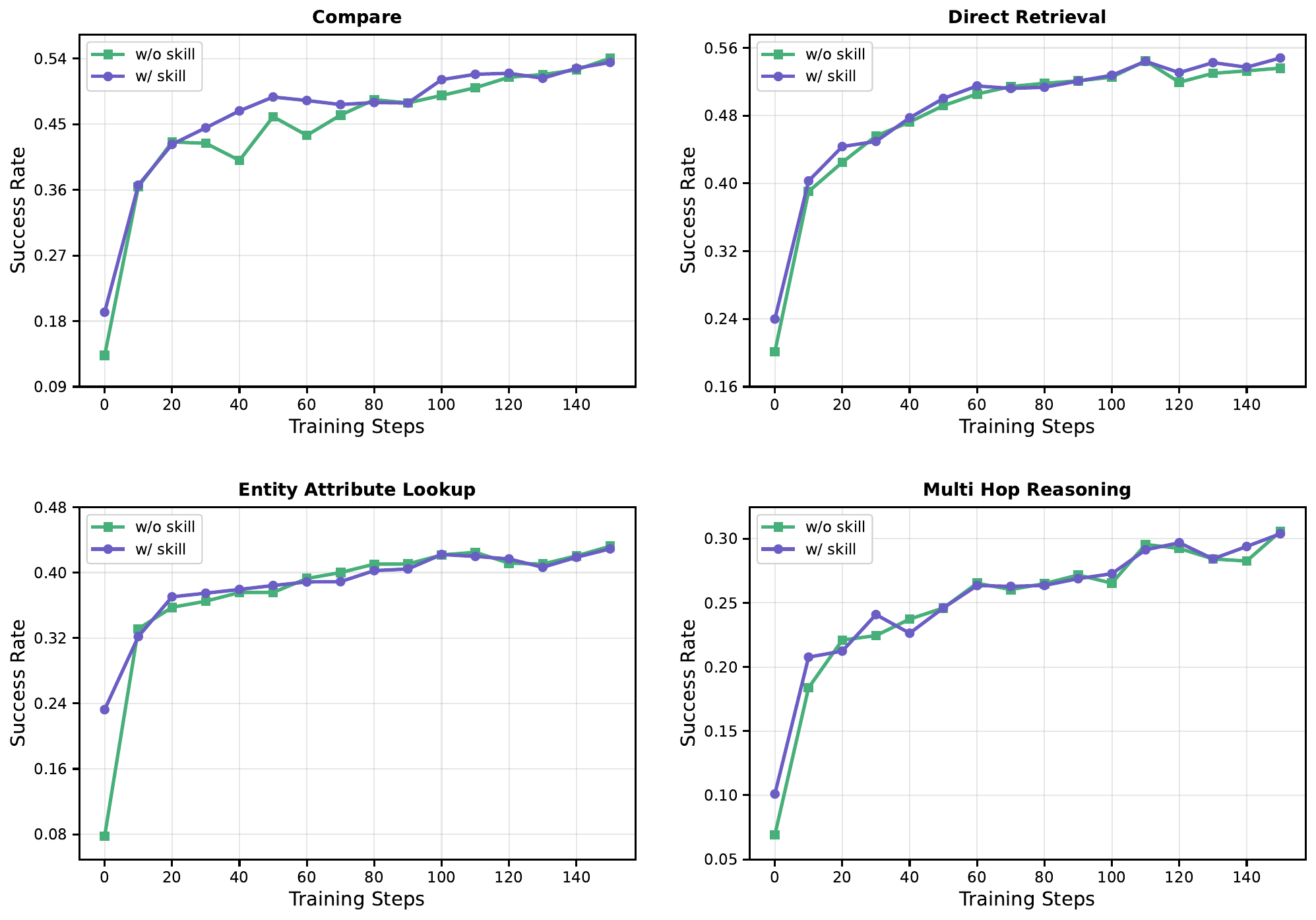}
\caption{Training dynamics of \methodname on Qwen2.5VL-3B, with SearchQA sub-tasks (split by skill categories) accuracy reported.}
\label{fig:subtask_comparison_searchqa}
\end{figure}

Figure~\ref{fig:subtask_comparison_alfworld} and Figure~\ref{fig:subtask_comparison_searchqa} present the per-subtask training dynamics of \methodname{} (Qwen2.5VL-3B) 
with and without skill context.
Across both ALFWorld and Search-QA, the \textit{w/ skill} result consistently 
achieves faster early-stage performance improvement.
The \textit{skill-free} result yields lower initial performance and gradually catches up toward the end of optimization, mirroring 
the skill internalization trend observed in Figure~\ref{fig:skill_comparison_1x3}.
These fine-grained per-subtask dynamics further confirm that the progressive annealing of skills 
drives the model to internalize task-relevant knowledge into its parameters.

\newpage
\section{Ablation Details}
We provide detailed ablation results of our curriculum design in Table~\ref{tab:detailed_ablations}, as a supplement for Figure~\ref{fig:ablations_skill_budget} and Table~\ref{tab:ablation_method}.
\begin{table*}[h]
\centering
\caption{\textbf{Detailed Results of Ablations.}}
\label{tab:detailed_ablations}
\resizebox{1\textwidth}{!}{%
\begin{tabular}{lcccccccc}
\toprule
\textbf{Method} & \textbf{Pick} & \textbf{Look} & \textbf{Clean} & \textbf{Heat} & \textbf{Cool} & \textbf{Pick2} & \textbf{Avg} & \textbf{vs. \methodname} \\
\midrule

\textbf{\methodname} (w/ $\mathcal{S}$)
    & 98.5 & 80.5 & 97.2 & 77.8 & 82.8 & 71.3 & 86.3 &  \\
\textbf{\methodname} (w/o $\mathcal{S}$)
    & 95.6 & 80.4 & 100  & 86.7 & 78.7 & 75.2 & 87.9 &  \\
$\Delta$
    & \negval{$\downarrow$2.9}
    & \negval{$\downarrow$0.1}
    & \posval{$\uparrow$2.8}
    & \posval{$\uparrow$8.9}
    & \negval{$\downarrow$4.1}
    & \posval{$\uparrow$3.9}
    & \posval{$\uparrow$1.6}
    & 
    \\
\midrule
\rowcolor{gray!10} \multicolumn{9}{l}{\textit{Ablations on Skill Budget}} \\

$[6,6,6]$ (w/ $\mathcal{S}$)
    & 90.4 & 69.4 & 97.0  & 95.2 & 74.1 & 81.3 & 85.9
    & \negval{$\downarrow$0.4} \\
$[6,6,6]$ (w/o $\mathcal{S}$)
    & 90.3 & 55.9 & 100.0 & 25.8 & 36.7 & 85.7 & 72.6
    & \negval{$\downarrow$15.3} \\
$\Delta$
    & \negval{$\downarrow$0.1}
    & \negval{$\downarrow$13.5}
    & \posval{$\uparrow$3.0}
    & \negval{$\downarrow$69.4}
    & \negval{$\downarrow$37.4}
    & \posval{$\uparrow$4.4}
    & \negval{$\downarrow$13.3}
    & 
    \\

$[6,4,2,1,0]$ (w/ $\mathcal{S}$)
    & 81.9 & 65.3 & 88.5 & 88.6 & 83.9 & 30.4 & 70.3
    & \negval{$\downarrow$16.0} \\
$[6,4,2,1,0]$ (w/o $\mathcal{S}$)
    & 80.4 & 66.8 & 87.9 & 93.7 & 58.5 & 40.7 & 71.1
    & \negval{$\downarrow$16.8} \\
$\Delta$
    & \negval{$\downarrow$1.5}
    & \posval{$\uparrow$1.5}
    & \negval{$\downarrow$0.6}
    & \posval{$\uparrow$5.1}
    & \negval{$\downarrow$25.4}
    & \posval{$\uparrow$10.3}
    & \posval{$\uparrow$0.8}
    & 
    \\
$[0,0,0]$ (w/o $\mathcal{S}$) & 95.7 & 67.8 & 80.1 & 63.6 & 83.2 &61.1 & 78.9 & \negval{$\downarrow$9.0} \\
\midrule
\rowcolor{gray!10} \multicolumn{9}{l}{\textit{Ablations on Dynamic Curriculum} (\texttt{Filter} \& \texttt{Rank} \& \texttt{Select})} \\

w/o \texttt{Filter} (w/ $\mathcal{S}$)
    & 91.7 & 65.9 & 97.5 & 73.0 & 73.0 & 74.1 & 81.6
    & \negval{$\downarrow$4.7} \\
w/o \texttt{Filter} (w/o $\mathcal{S}$)
    & 91.0 & 45.0 & 98.3 & 65.5 & 71.9 & 71.5 & 78.9
    & \negval{$\downarrow$9.0} \\
$\Delta$
    & \negval{$\downarrow$0.7}
    & \negval{$\downarrow$20.9}
    & \posval{$\uparrow$0.8}
    & \negval{$\downarrow$7.5}
    & \negval{$\downarrow$1.1}
    & \negval{$\downarrow$2.6}
    & \negval{$\downarrow$2.7}
    & 
    \\

w/o \texttt{Rank} (w/ $\mathcal{S}$)
    & 92.0 & 62.4 & 93.4 & 86.7 & 46.1 & 68.4 & 76.6
    & \negval{$\downarrow$9.7} \\
w/o \texttt{Rank} (w/o $\mathcal{S}$)
    & 88.4 & 42.3 & 95.5 & 25.0 & 25.3 & 56.5 & 62.9
    & \negval{$\downarrow$25.0} \\
$\Delta$
    & \negval{$\downarrow$3.6}
    & \negval{$\downarrow$20.1}
    & \posval{$\uparrow$2.1}
    & \negval{$\downarrow$61.7}
    & \negval{$\downarrow$20.8}
    & \negval{$\downarrow$11.9}
    & \negval{$\downarrow$13.7}
    & 
    \\

\bottomrule
\end{tabular}%
}
\end{table*}
\newpage

\section{More Comparisons}

\begin{table*}[h]
\centering
\caption{\textbf{Comparison on ALFWorld benchmark}. $^*$ denotes the results trained with GRPO.}
\label{tab:alfworld}
\resizebox{0.9\textwidth}{!}{%
\begin{tabular}{lccccccc}
\toprule
\textbf{Method} & \textbf{Pick} & \textbf{Look} & \textbf{Clean} & \textbf{Heat} & \textbf{Cool} & \textbf{Pick2} & \textbf{Avg} \\
\midrule
\rowcolor{gray!10} \multicolumn{8}{l}{\textit{Closed-source LLMs}} \\
GPT-4o~\citep{hurst2024gpt-4o}          & 75.3 & 60.8 & 31.2 & 56.7 & 21.6 & 49.8 & 48.0 \\
Gemini-2.5-Pro~\citep{comanici2025gemini}  & 92.8 & 63.3 & 62.1 & 69.0 & 26.6 & 58.7 & 60.3 \\
\midrule
\rowcolor{gray!10} \multicolumn{8}{l}{\textit{Qwen2.5-(VL)-3B-Instruct}} \\
Vanilla~\citep{bai2023qwen}         & 27.0 &24.3& 4.5 &20.5& 10.2& 0.0 &15.2 \\
GRPO~\citep{shao2024deepseekmath} &92.6 &85.7& 70.6& 86.6 &79.3 &65.0 &79.9 \\
EvolveR~\citep{wu2025evolver} &77.3 &24.5& 47.9& 41.7& 24.6 &22.5 &44.1 \\
AgentOCR~\citep{feng2026agentocr} &91.9& 81.8& 76.0& 73.3& 76.1& 70.0 &78.2 \\
\textbf{\methodname (Ours)} & 95.6 &80.4& 100& 86.7& 78.7& 75.2& 87.9\\
\midrule
\rowcolor{gray!10} \multicolumn{8}{l}{\textit{Qwen2.5-(VL)-7B-Instruct}} \\
Vanilla~\citep{bai2023qwen}         & 33.4 & 21.6 & 19.3 &  6.90 &  2.80 &  3.20 & 14.8 \\
ReAct~\citep{yao2022react}       & 48.5 & 35.4 & 34.3 & 13.2 & 18.2 & 17.6 & 31.2 \\
Reflexion~\citep{shinn2024reflexion} & 62.0 & 41.6 & 44.9 & 30.9 & 36.3 & 23.8 & 42.7 \\
Mem0~\citep{chhikara2025mem0}            & 54.0 & 55.0 & 26.9 & 36.4 & 20.8 &  7.69 & 33.6 \\
ExpeL~\citep{zhao2024expel}           & 21.0 & 67.0 & 55.0 & 52.0 & 71.0 &  6.00 & 46.3 \\
MemP~\citep{fang2025memp}            & 54.3 & 38.5 & 48.1 & 56.2 & 32.0 & 16.7 & 41.4 \\
SimpleMem~\citep{liu2026simplemem}       & 64.5 & 33.3 & 20.0 & 12.5 & 33.3 &  3.84 & 29.7 \\
RLOO~\citep{ahmadian2024rloo}        & 87.6 & 78.2 & 87.3 & 81.3 & 71.9 & 48.9 & 75.5 \\
GRPO~\citep{shao2024deepseekmath}        & 90.8 & 66.1 & 89.3 & 74.7 & 72.5 & 64.7 & 77.6 \\

MemRL~\citep{zhang2026memrl}           & 62.8 & 38.5 & 22.2 & 12.5 &  8.00 &  0.00 & 21.4 \\

EvolveR~\citep{wu2025evolver}         & 64.9 & 33.3 & 46.4 & 13.3 & 33.3 & 33.3 & 43.8 \\
Mem0$^*$~\citep{chhikara2025mem0}      & 78.1 & 54.8 & 56.1 & 31.0 & 65.0 & 26.9 & 54.7 \\
SimpleMem$^*$~\citep{liu2026simplemem}  & 89.5 & 36.3 & 60.0 & 50.0 & 64.9 & 26.3 & 62.5 \\
AgentOCR~\citep{feng2026agentocr} &95.6 &96.2 &78.1 &73.2& 72.4 &72.0 &81.2 \\
\textbf{\methodname (Ours)}&100 & 85.8 & 94.6 & 81.9 & 85.7 & 80.1 & 89.8 \\
\bottomrule
\end{tabular}%
}
\end{table*}

Table~\ref{tab:alfworld} and Table~\ref{tab:searchqa} present extended comparisons 
against a broader set of baselines beyond those reported in Table~\ref{tab:main_results}.
On ALFWorld, \methodname{} achieves average success rates of 87.9 (3B) and 89.8 (7B), 
substantially outperforming memory-augmented approaches such as 
ExpeL (46.3), SimpleMem (62.5), Mem0 (54.7), and MemRL (21.4), 
as well as closed-source models including GPT-4o (48.0) and Gemini-2.5-Pro (60.3).
On Search-QA, \methodname{} attains average scores of 40.8 (3B) and 44.4 (7B), 
surpassing retrieval-augmented and search-based methods including 
RAG (27.0/30.4), Search-R1 (32.5/38.5), ZeroSearch (31.7/39.1), and EvolveR (38.2/43.1).
Notably, \methodname{} achieves particularly strong performance on out-of-domain 
multi-hop datasets such as Bamboogle (63.7/66.9), 
highlighting its robust generalization to unseen reasoning tasks 
without any domain-specific adaptation.
\begin{table*}[h]
\centering
\caption{\textbf{Results on Search-based QA.} $^\dagger$ and $^\star$ denote in-domain and out-of-domain respectively.}
\label{tab:searchqa}
\resizebox{\textwidth}{!}{%
\begin{tabular}{lcccccccc}
\toprule
\multirow{2}{*}{\textbf{Method}} 
    & \multicolumn{3}{c}{\textbf{Single-Hop QA}} 
    & \multicolumn{4}{c}{\textbf{Multi-Hop QA}} 
    & \multirow{2}{*}{\textbf{Avg.}} \\
\cmidrule(lr){2-4} \cmidrule(lr){5-8}
& \textbf{NQ}$^\dagger$ 
& \textbf{TriviaQA}$^\star$ 
& \textbf{PopQA}$^\star$ 
& \textbf{HotpotQA}$^\dagger$ 
& \textbf{2Wiki}$^\star$ 
& \textbf{MuSiQue}$^\star$ 
& \textbf{Bamboogle}$^\star$ & \\

\midrule
\rowcolor{gray!10} \multicolumn{9}{l}{\textit{Qwen2.5-(VL)-3B-Instruct}} \\
Vanilla          & 12.4 & 30.6 &  5.6 & 16.0 & 19.2 &  4.4 & 16.8 & 15.0 \\
CoT              & 15.0 & 33.6 &  3.6 & 16.2 & 18.0 &  3.6 & 12.8 & 14.7 \\
RAG              & 34.8 & 54.4 & 38.7 & 25.5 & 22.6 &  4.7 &  8.0 & 27.0 \\
RA-Agent &15.2& 28.4& 6.6& 12.6 &16.6& 2.6& 13.6& 13.7 \\

IRCoT      & 11.1 & 31.2 & 20.0 & 16.4 & 17.1 &  6.7 & 24.0 & 18.1 \\
Search-o1       & 16.6 & 31.0 &  8.2 & 14.8 & 22.4 &  5.2 & 22.4 & 17.2 \\
SFT        & 24.9 & 29.2 & 10.4 & 18.6 & 24.8 &  4.4 & 11.2 & 17.6 \\
R1-Instruct& 21.0 & 44.9 & 17.1 & 20.8 & 27.5 &  6.0 & 19.2 & 22.4 \\
Reject Sampling & 29.4 & 48.8 & 33.2 & 24.0 & 23.3 & 5.9 & 21.0 & 26.5 \\
Search-R1  & 34.1 & 54.5 & 37.8 & 32.4 & 31.9 & 10.3 & 26.4 & 32.5 \\
ZeroSearch       & 41.4 & 57.4 & 44.8 & 27.4 & 30.0 &  9.8 & 11.1 & 31.7 \\
StepSearch &- & - & - & 34.5 & 32.0 & 17.4 & 34.4& -- \\
EvolveR &43.4 &58.4 &43.4 &37.3 &38.1 &13.7& 32.8 &38.2\\
\textbf{\methodname (Ours)} & 39.8 &57.5& 42.3& 35.1& 33.7 &13.3& 63.7& 40.8\\
\midrule
\rowcolor{gray!10} \multicolumn{9}{l}{\textit{Qwen2.5-(VL)-7B-Instruct}} \\
Vanilla          & 11.6 & 35.6 &  1.2 & 16.4 & 22.2 &  4.8 & 14.4 & 15.2 \\
CoT              & 12.8 & 35.6 &  3.8 & 16.2 & 22.6 &  6.6 & 24.0 & 17.4 \\
RAG              & 34.9 & 58.5 & 39.2 & 29.9 & 23.5 &  5.8 & 20.8 & 30.4 \\
RA-Agent &21.2& 40.2& 8.8 &19.6 &19.6& 7.6 &28.0& 20.7 \\
IRCoT      & 22.4 & 47.8 & 30.1 & 13.3 & 14.9 &  7.2 & 22.4 & 23.9 \\
Search-o1        & 19.4 & 40.6 & 11.4 & 17.0 & 27.0 &  8.6 & 30.4 & 22.1 \\
SFT        & 31.8 & 35.4 & 12.1 & 21.7 & 25.9 &  6.6 & 11.2 & 20.7 \\
R1-Instruct& 27.0 & 53.7 & 19.9 & 23.7 & 29.2 &  7.2 & 29.3 & 27.1 \\
Reject Sampling & 36.0 & 59.2 & 38.0 & 33.1 & 29.6 & 12.3 & 35.5 & 34.8 \\
Search-R1  & 39.3 & 61.0 & 39.7 & 37.0 & 41.4 & 14.6 & 36.8 & 38.5 \\
ZeroSearch       & 43.6 & 61.8 & 51.5 & 34.6 & 35.2 & 18.4 & 27.8 & 39.1 \\
StepSearch           & --   & --   & --   & 38.6 & 36.6 & 22.6 & 40.0 & --   \\
EvolveR              & 43.5 & 63.4 & 44.6 & 38.2 & 42.0 & 15.6 & 54.4 & 43.1 \\
\textbf{\methodname (Ours)} 
    & 42.7& 61.1& 45.3 &40.0 &38.3& 16.4& 66.9& 44.4  \\
\bottomrule
\end{tabular}%
}
\end{table*}

\section{Implementation Details}

We follow the rendering configurations in \citet{feng2026agentocr} to construct the visual context 
in \methodname{} for each benchmark, with the full prompts shown in Figure~\ref{fig:prompt_alfworld} 
and Figure~\ref{fig:prompt_searchqa}.
Text is rendered in a monospace font with a line spacing of 1.2 across both environments, 
with a font size of 10pt and a maximum width of 392px for ALFWorld and WebShop, 
and 12pt with 560px for Search-QA.
To enable visual disambiguation of different context components, 
we apply a semantic color coding scheme: task instructions and general context 
are rendered in black, while for ALFWorld, observations are highlighted in 
\textcolor[RGB]{0,0,255}{blue} and actions in \textcolor[RGB]{255,0,0}{red}; 
for Search-QA, the same convention is applied to \texttt{<search>} queries 
and \texttt{<information>} results, respectively, 
allowing the vision encoder to clearly distinguish between 
perceived states, executed actions, and retrieved content at a glance. Table~\ref{tab:skillbank} and Table~\ref{tab:skillbank_webshop} present representative examples of the skill files 
stored in \texttt{SkillBank}, illustrating the structured procedural knowledge 
provided to the agent across all three task categories.

\begin{table*}[h]
\centering
\caption{\textbf{Representative Skills in \texttt{SkillBank}.}}
\label{tab:skillbank}
\resizebox{\textwidth}{!}{%
\begin{tabular}{p{3.8cm} p{7.5cm} p{5.0cm}}
\toprule
\textbf{Skill Title} & \textbf{Principle (Actionable Pattern)} & \textbf{When to Apply} \\
\midrule

\rowcolor{gray!10}
\multicolumn{3}{l}{\texttt{skills/ALFWorld/general.md}} \\

Systematic Exploration 
    & Search every plausible surface or container exactly once before revisiting; prioritize unseen locations. 
    & Anytime the goal count is not met and unexplored areas remain. \\

Immediate Acquisition 
    & As soon as a required object becomes visible and reachable, take it immediately before moving elsewhere. 
    & Upon first visual confirmation of a goal-relevant object. \\

\midrule
\rowcolor{gray!10}
\multicolumn{3}{l}{\texttt{skills/ALFWorld/pick\_and\_place.md}} \\

Grab When Seen 
    & Whenever a needed object is visible and reachable, immediately take it before moving elsewhere. 
    & Upon first sight of an unheld object matching the goal specification. \\

Place Before More Search 
    & When holding a goal object and the target location is known, navigate there and place it immediately. 
    & While carrying a required object and the destination has been identified. \\

\midrule
\rowcolor{gray!10}
\multicolumn{3}{l}{\texttt{skills/ALFWorld/look\_at\_obj\_in\_light.md}} \\

Switch Lamp On 
    & Issue the \texttt{use desklamp} command as soon as you reach it so the light condition is satisfied. 
    & Upon arriving at a desklamp that is currently off. \\

Grab Target First 
    & If the target is visible but the desklamp is not, take the target immediately to carry it to the lamp. 
    & When the target is visible and not yet held, while desklamp location is unknown. \\

\midrule
\rowcolor{gray!10}
\multicolumn{3}{l}{\texttt{skills/ALFWorld/clean.md}} \\

Phase-Ordered Plan 
    & Execute in fixed sequence: (1) locate \& acquire, (2) clean at sink, (3) navigate, (4) place. 
    & As soon as the goal specifies the object must be clean before placement. \\

Sink First for Cleaning 
    & Upon holding the target, go straight to the nearest sink and issue the clean command. 
    & Once the target is in hand and its required state is clean. \\

\midrule
\rowcolor{gray!10}
\multicolumn{3}{l}{\texttt{skills/ALFWorld/heat.md}} \\

Secure Exact Target First 
    & Identify and pick up the exact object named in the goal before interacting with the microwave. 
    & After spotting any candidate object, before opening or using appliances. \\

Open Then Heat 
    & Upon reaching the microwave with the target in hand, open the door, place the object, then heat. 
    & Immediately after navigating to the microwave with the target object held. \\

\midrule
\rowcolor{gray!10}
\multicolumn{3}{l}{\texttt{skills/ALFWorld/cool.md}} \\

Prep Cooling Appliance 
    & Locate the fridge first and open it so it is ready before or immediately after grabbing the target. 
    & As soon as the fridge comes into view or right after acquiring the target object. \\

Enforce Cooling Before Placement 
    & Do not place the target object in its final location until a cooling action has been successfully executed. 
    & When holding the correct object and before any placement action is attempted. \\

\midrule
\rowcolor{gray!10}
\multicolumn{3}{l}{\texttt{skills/Search/general.md}} \\

Decompose Then Search
    & Break the question into minimal sub-questions and handle each with its own targeted query before synthesizing.
    & Any complex or multi-hop question requiring multiple intermediate facts. \\

Exit When Evidence Is Solid
    & Stop issuing further queries once clear, corroborated evidence is found; avoid premature termination.
    & After each read step---answer only if confidence is justified, otherwise refine search. \\

\midrule
\rowcolor{gray!10}
\multicolumn{3}{l}{\texttt{skills/Search/direct\_retrieval.md}} \\

Isolate Core Query
    & Strip the question to its key entity plus sought fact and search exactly that pair first.
    & At the start of any direct-retrieval task. \\

Evidence-Bound Answer
    & Only state an answer explicitly supported by retrieved text; continue searching rather than guess.
    & Before finalizing any factoid answer. \\

\midrule
\rowcolor{gray!10}
\multicolumn{3}{l}{\texttt{skills/Search/multi\_hop\_reasoning.md}} \\

Targeted Sequential Searches
    & Issue separate, focused searches for each sub-question instead of one broad query.
    & After decomposition, when distinct pieces of information must be collected individually. \\

Collect-Then-Compare
    & Retrieve concrete values for all items before performing any comparison or conclusion.
    & For comparative tasks involving dates, places, or quantitative attributes. \\

\midrule
\rowcolor{gray!10}
\multicolumn{3}{l}{\texttt{skills/Search/entity\_attribute\_lookup.md}} \\

Direct Attribute Query
    & Include both the full entity name and target attribute in the first search to surface authoritative results.
    & Whenever the entity's full, unambiguous name is provided in the question. \\

Two-Source Cross-Check
    & Confirm the attribute in at least two independent, reputable sources to avoid hallucinations.
    & After the first plausible answer appears or when the attribute seems uncommon or uncertain. \\

\midrule
\rowcolor{gray!10}
\multicolumn{3}{l}{\texttt{skills/Search/compare.md}} \\

Parallel Attribute Lookup
    & Independently retrieve the identical attribute for each entity via separate, focused searches.
    & After identifying entities and the comparison attribute. \\

Normalize Before Comparing
    & Convert retrieved values to a common comparable form before judging equality or ordering.
    & After gathering each entity's attribute but before drawing any conclusion. \\

\bottomrule
\end{tabular}%
}
\end{table*}

\begin{table*}[h]
\centering
\caption{\textbf{Representative Skills in \texttt{SkillBank} for WebShop.}}
\label{tab:skillbank_webshop}
\resizebox{\textwidth}{!}{%
\begin{tabular}{p{3.8cm} p{7.5cm} p{5.0cm}}
\toprule
\textbf{Skill Title} & \textbf{Principle (Actionable Pattern)} & \textbf{When to Apply} \\
\midrule

\rowcolor{gray!10}
\multicolumn{3}{l}{\texttt{skills/WebShop/general.md}} \\

Prioritize Core Keywords
    & Include product type, 1--2 key functional attributes, and hard constraints
      (price, size, color) in the search query; omit secondary descriptors to avoid
      over-constraining.
    & Before issuing the first search or when refining an over-specific query that
      yields few results. \\

Scan Before You Click
    & Read product titles, thumbnails, and prices in the results list to ensure the
      item plausibly meets core constraints before opening it.
    & On any search results page when deciding which product link to open next. \\

Set Mandatory Variants
    & Always select required variant options (size, color, capacity) before
      evaluating price or purchasing.
    & After confirming the product type matches and before any purchase action. \\

Purchase Decisively
    & Once all constraints are confirmed on a variant and price fits, execute Buy
      Now without unnecessary further navigation.
    & After validating every constraint on the current product variant. \\

\midrule
\rowcolor{gray!10}
\multicolumn{3}{l}{\texttt{skills/WebShop/apparel.md}} \\

Focus Key Query
    & Include only product type plus must-have attributes (gender, garment, fabric,
      fit, price cap) in the search; drop minor terms to widen relevant results.
    & Before issuing or refining any search query for apparel items. \\

Check Variant Price
    & After choosing size and color, verify the displayed price for that specific
      variant is within budget; abandon if it exceeds the limit.
    & Immediately after variant selection and before clicking Buy Now. \\

\midrule
\rowcolor{gray!10}
\multicolumn{3}{l}{\texttt{skills/WebShop/electronics.md}} \\

Constraint-Rich Search
    & Pack product type plus every mandatory attribute (features, color, size, price
      cap) into the initial search string to surface only highly relevant electronics
      items.
    & When starting a new product hunt or refining after poor results. \\

Bail on Mismatch Fast
    & If variant clicks do not update product details to the required spec, use Back
      and seek another item instead of retrying the same option.
    & When repeated option clicks leave title or specs unchanged or incompatible. \\

\midrule
\rowcolor{gray!10}
\multicolumn{3}{l}{\texttt{skills/WebShop/footwear.md}} \\

Verify Key Features
    & Open the product description or specs to explicitly confirm required functional
      attributes (e.g., slip-resistant, rubber sole) before purchasing.
    & Immediately after opening a product page, before selecting variants or buying. \\

Exit Non-Matches Fast
    & If a product clearly violates any hard constraint, back out immediately instead
      of toggling options or repeating identical searches.
    & Upon noticing missing attributes, wrong category, unavailable size, or
      over-budget price. \\

\midrule
\rowcolor{gray!10}
\multicolumn{3}{l}{\texttt{skills/WebShop/beauty\_health.md}} \\

Feature-Led Click
    & Open the first result whose title explicitly mentions the key functional
      attribute to maximize match likelihood.
    & After search results appear and at least one headline contains the core
      feature term. \\

Minimal Path Purchase
    & Once a product fully satisfies all constraints, proceed directly to Buy Now
      without extra browsing to reduce error risk.
    & When all user requirements are confirmed on the current product page. \\

\midrule
\rowcolor{gray!10}
\multicolumn{3}{l}{\texttt{skills/WebShop/home\_decor.md}} \\

Use Variant Selectors
    & Systematically choose color, size, shape, and other visible variants,
      confirming each selection updates the listing to match constraints and price.
    & After landing on a suitable product that offers configurable attributes. \\

Open Details For Hidden Specs
    & Expand Description or Features tabs to confirm less-visible requirements
      (material, washability, printing method) not obvious from titles.
    & When any user constraint is not directly visible in the main listing or
      variant options. \\

\midrule
\rowcolor{gray!10}
\multicolumn{3}{l}{\texttt{skills/WebShop/accessories.md}} \\

Explicit Variant Selection
    & Always click and visibly confirm the exact color, size, or pattern variant
      instead of assuming the default matches the request.
    & After opening a product that offers multiple variant options. \\

Post-Config Price Audit
    & After selecting all variants, re-confirm the displayed price is within budget;
      if exceeded, backtrack and search for alternatives.
    & Just before initiating checkout or adding to cart. \\

\bottomrule
\end{tabular}%
}
\end{table*}
\begin{figure}[h]
\centering
\begin{templatebox}{Prompt of \methodname on ALFWorld}
You are an expert agent operating in the ALFRED embodied Environment.
\{skill\_context\} Your task is to: \{task\_description\}.
Prior to this step, you have already taken \{step\_count\} step(s).
The provided image shows the most recent \{history\_length\} observations and the corresponding actions you took.
You are now at step \{current\_step\} and your current observation is: \{current\_observation\}.
Your admissible actions of the current situation are: [\{admissible\_actions\}].

Now it's your turn to take an action.
You should first reason step-by-step about the current situation.
This reasoning process \textbf{MUST} be enclosed within \texttt{<think> </think>} tags.
Once you've finished your reasoning, you should choose an admissible action for current step and present it within \texttt{<action> </action>} tags.

Additionally, select an image compression factor larger than 1.0 for the next image.
Higher compression lowers cost, but too much compression harms image quality.
You must provide the next compression factor within \texttt{<compression> </compression>} tags (e.g., \texttt{<compression>1.1</compression>}).
\end{templatebox}
\caption{Prompt template used by \methodname{} for the ALFWorld embodied task environment.}
\label{fig:prompt_alfworld}
\end{figure}

\begin{figure}[h]
\centering
\begin{templatebox}{Prompt of \methodname on Search-based QA}
You are an expert agent tasked with answering the given question step-by-step.
\{skill\_context\}
Your question: \{task\_description\}.
Prior to this step, you have already taken \{step\_count\} step(s).
The image contains the full history:
\begin{itemize}
    \item Past queries are inside \texttt{<search>...</search>}
    \item Past results are inside \texttt{<information>...</information>}
\end{itemize}
Now it's your turn to respond for the current step.
You should first conduct a reasoning process.
After completing your reasoning, choose only one of the following actions (do not perform both):
\begin{enumerate}
    \item If any required knowledge is missing or uncertain, you \textbf{MUST} call a search engine to get more external information using format: \texttt{<search> your query </search>}.
    \item Only if you have sufficient information to answer the question with high confidence, provide your final answer within \texttt{<answer> </answer>} tags.
\end{enumerate}
Additionally, select an image compression factor larger than 1.0 for the next image.
Higher compression lowers cost, but too much compression harms image quality.
You must provide the next compression factor within \texttt{<compression> </compression>} tags (e.g., \texttt{<compression>1.1</compression>}).

\textbf{Output format:}
\begin{enumerate}
    \item Reasoning: state what you found in the image.
    \item \texttt{<search>...</search>} or \texttt{<answer>...</answer>}
    \item \texttt{<compression>...</compression>}
\end{enumerate}
\end{templatebox}
\caption{Prompt template used by \methodname{} for the Search-based QA task environment.}
\label{fig:prompt_searchqa}
\end{figure}
 \begin{figure}[h]                                                                   
  \centering                                                                          
  \begin{templatebox}{Prompt of \methodname on WebShop}                               
  You are an expert autonomous agent operating in the WebShop e‑commerce environment. 
  \{skill\_context\}                                                                  
  Your task is to: \{task\_description\}.                                             
  Prior to this step, you have already taken \{step\_count\} step(s). Below are the   
  most recent \{history\_length\} observations and the corresponding actions you took:
   \{action\_history\}                                                                
  You are now at step \{current\_step\} and your current observation is:              
  \{current\_observation\}.                                                           
  Your admissible actions of the current situation are:
  [                                                                                   
  \{available\_actions\}                                                              
  ].                                                                                  
  The image contains the full history:                                                
  \begin{itemize}                                                                     
      \item Past observations are the page content after each action                  
      \item Past actions are inside \texttt{<action>...</action>}                     
  \end{itemize}                                                                       
  Now it's your turn to take one action for the current step.                         
  You should first reason step-by-step about the current situation, then think        
  carefully which admissible action best advances the shopping goal. This reasoning   
  process MUST be enclosed within \texttt{<think> </think>} tags.                     
  Once you've finished your reasoning, you should choose an admissible action for the 
  current step and present it within \texttt{<action> </action>} tags.                
   
  Additionally, select an image compression factor larger than 1.0 for the next image.
  Higher compression lowers cost, but too much compression harms image quality.
  You must provide the next compression factor within \texttt{<compression>           
  </compression>} tags (e.g., \texttt{<compression>1.1</compression>}).               
                                                                                      
  \textbf{Output format:}                                                             
  \begin{enumerate}
      \item \texttt{<think>}Reasoning: state what you found in the image and which    
  action best advances the goal.\texttt{</think>}                                     
      \item \texttt{<action>...</action>}                                             
      \item \texttt{<compression>...</compression>}                                   
  \end{enumerate}                                                                     
  \end{templatebox}                                                                   
  \caption{Prompt template used by \methodname{} for the WebShop task environment.}   
  \label{fig:prompt_webshop}                                                          
  \end{figure}


\end{document}